\documentclass[journal]{IEEEtran}

\usepackage{amsmath,amsfonts,amssymb}
\usepackage{algorithm}
\usepackage{algorithmic}
\usepackage{array}
\usepackage{tabularx}
\usepackage[caption=false,font=footnotesize]{subfig}
\usepackage{textcomp}
\usepackage{stfloats}
\usepackage{url}
\usepackage{graphicx}
\usepackage{cite}
\usepackage{booktabs}
\usepackage{multirow}
\usepackage[table]{xcolor}
\usepackage{xspace}
\usepackage{makecell}
\usepackage{mathtools}
\usepackage{placeins}

\newcommand{\View}{\mathcal{M}}
\newcommand{\Araw}{\mathbf{A}}
\newcommand{\Ahat}{\widehat{\mathbf{A}}}
\newcommand{\Atilde}{\widetilde{\mathbf{A}}}
\newcommand{\WU}{\mathbf{W}_{U}}
\newcommand{\zv}{\mathbf{z}}
\newcommand{\htxt}{\mathbf{h}}
\newcommand{\rank}{\operatorname{rank}}
\newcommand{\softmax}{\operatorname{softmax}}

\newcommand{\RBO}{\operatorname{RBO}}
\newcommand{\warp}{\Omega}
\newcommand{\pospart}[1]{\left\lfloor #1 \right\rfloor_{+}}
\DeclareMathOperator*{\argmin}{arg\,min}

\newcolumntype{Y}{>{\centering\arraybackslash}X}

\definecolor{lightgray}{gray}{0.92}

\newcommand{\method}{ERCR\xspace}
\newcommand{\er}{ER\xspace}
\newcommand{\pcr}{PCR\xspace}

\begin{document}

\title{Evidence Recomposition and Predictive Context Residualization for Visual Attribution in Multimodal Large Language Models}

\author{Jiawei~Liang, Jianjie~Huang, Xianghao~Jiao,
Siyuan~Liang, \\ Shiming~Liu, and Xiaochun~Cao,~\IEEEmembership{Senior Member,~IEEE}%
\thanks{Jiawei Liang and Jianjie Huang are with Shenzhen Campus of Sun Yat-sen University, Shenzhen 518107, China, and also with Zhongguancun Academy, Beijing 100094, China (e-mail: liangjw57@mail2.sysu.edu.cn; huangjj67@mail2.sysu.edu.cn).}%
\thanks{Xianghao Jiao and Xiaochun Cao are with Shenzhen Campus of Sun Yat-sen University, Shenzhen 518107, China (e-mail: jiaoxh0331@outlook.com; caoxiaochun@mail.sysu.edu.cn).}%
\thanks{Siyuan Liang is with Nanyang Technological University, Singapore (e-mail: siyuan.liang@ntu.edu.sg).}%
\thanks{Shiming Liu is with the Department of Mechanical Engineering, Imperial College London, London, U.K. (e-mail: 852074479@qq.com).}%
\thanks{Corresponding author: Xiaochun Cao.}
}

\markboth{IEEE Transactions on Image Processing,~Vol.~XX, No.~XX, 2026}%
{Liang \MakeLowercase{\textit{et al.}}: Evidence Recomposition and Predictive Context Residualization}

\maketitle

\begin{abstract}
Multimodal large language models (MLLMs) have achieved strong vision-language performance, yet their token-level visual evidence remains difficult to inspect. Recent logit-lens attribution methods project each visual-token hidden state into the vocabulary space to explain generated words, but this token-wise readout introduces a mismatch: visual tokens are context-mixed by the model, while the attribution score is decoded independently at each token location. This often produces fragmented attribution maps and can be further affected by autoregressive context signals from preceding text tokens. We propose \method, an attribution framework built from Evidence Recomposition (ER) and Predictive Context Residualization (PCR). ER aggregates target evidence across multiple views with different token-to-region assignments, reducing attribution fragmentation caused by a single readout grid. PCR estimates a preceding-token context map with RBO-based rank relevance and subtracts its fitted component from the ER map to suppress context-token interference. Experiments on LLaVA, Qwen2-VL, and InternVL families across COCO Caption, GranDf, and OpenPSG show that \method improves visual evidence for target tokens and mitigates preceding-token context interference under the existing evaluation protocol. On Qwen2-VL-2B, \method improves TAM F1-IoU from 39.10 to 44.45 on COCO Caption and from 30.83 to 37.20 on GranDf. Overall, \method provides a practical refinement for token-level visual evidence inspection.

\end{abstract}


\section{Introduction}
\label{sec:introduction}

\IEEEPARstart{M}{ultimodal} Large Language Models (MLLMs)~\cite{liu2024visual,wang2024qwen2,chen2024expanding,liang2025revisiting,liang2025vl} have become a common interface for image captioning, visual question answering, and open-ended visual reasoning. Their outputs are easy to observe, but the visual evidence behind each generated word is still difficult to verify. This gap limits reliable deployment~\cite{kong2025universal,liu2024efficient} in settings where users need to diagnose hallucinations, distinguish visually grounded statements from language priors, or understand why a model refers to a particular object or relation.

\begin{figure}[!t]
    \centering
    \includegraphics[width=\linewidth]{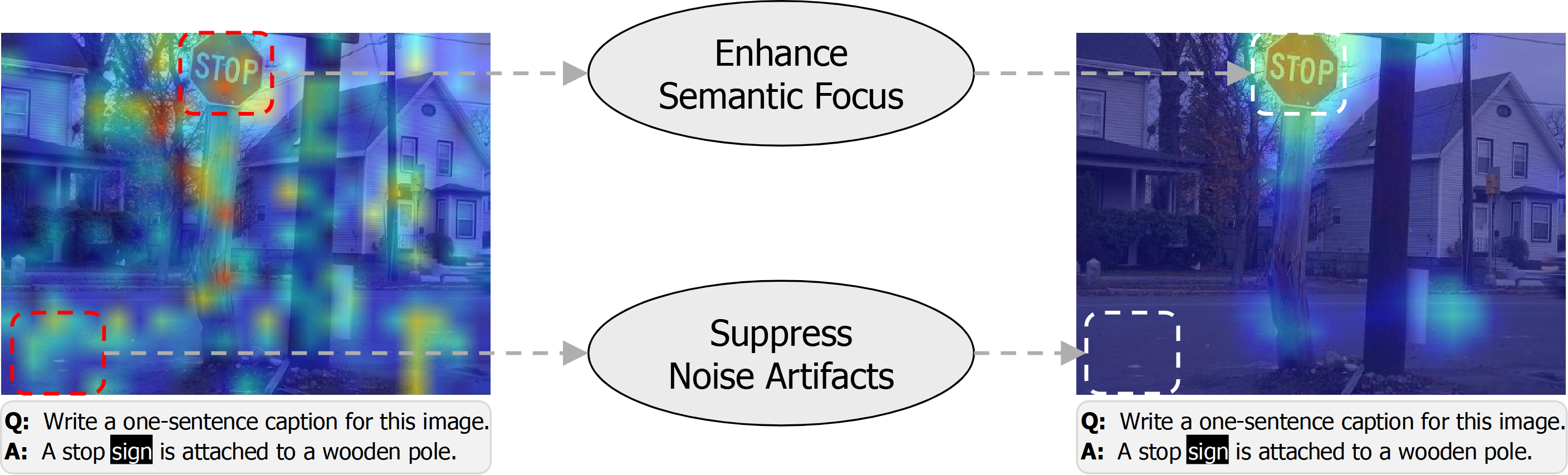}
    \caption{Our goal is to estimate token-level visual evidence with less dependence on a single readout grid. \method aggregates evidence across multiple views and mitigates context interference, yielding attribution maps that are more focused on target-relevant regions.}
    \label{fig:main}
\end{figure}

Visual attribution methods for MLLMs commonly adapt gradients, attention propagation, or logit-lens readouts to identify which image regions support a generated token~\cite{ben2024lvlm,zhang2024redundancy,jiang2024interpreting,li2025token}. Among these, logit-lens attribution directly decodes visual-token hidden states through the language unembedding matrix. Given a target output token, each visual token is assigned the target-token logit obtained from its hidden state, providing a simple token-level attribution interface for autoregressive MLLMs.

A useful MLLM attribution method should produce spatially coherent support for generated object words, suppress low-relevance visual evidence, and remain stable under routine preprocessing choices such as image resizing and canvas padding. These requirements connect token-level MLLM attribution to saliency-map evaluation, weakly supervised localization, and perturbation-based visual evidence tests.

However, token-wise logit-lens attribution has two practical instability sources: single-grid readout fragmentation and preceding-token context interference. First, visual tokens are not independent local patches after MLLM processing; they contain context-mixed information from neighboring and global regions. Independently unembedding each token location can therefore assign broader regional evidence to a single visual tokenization grid, producing fragmented maps with limited object support. Second, the target token $T_t$ is predicted from the image and preceding text tokens $T_{<t}$. Under this autoregressive prediction context, the visual-token readout for $k_t$ can produce a map that shares spatial patterns with attribution maps of preceding tokens, including earlier words, punctuation, or connectors. We refer to this overlap as preceding-token context interference. Such overlap can introduce diffuse activation or activation unrelated to the current target.

\begin{figure*}[!t]
    \centering
    \subfloat[Token-to-region assignments affect logit-lens readouts.]{
        \includegraphics[width=0.43\linewidth]{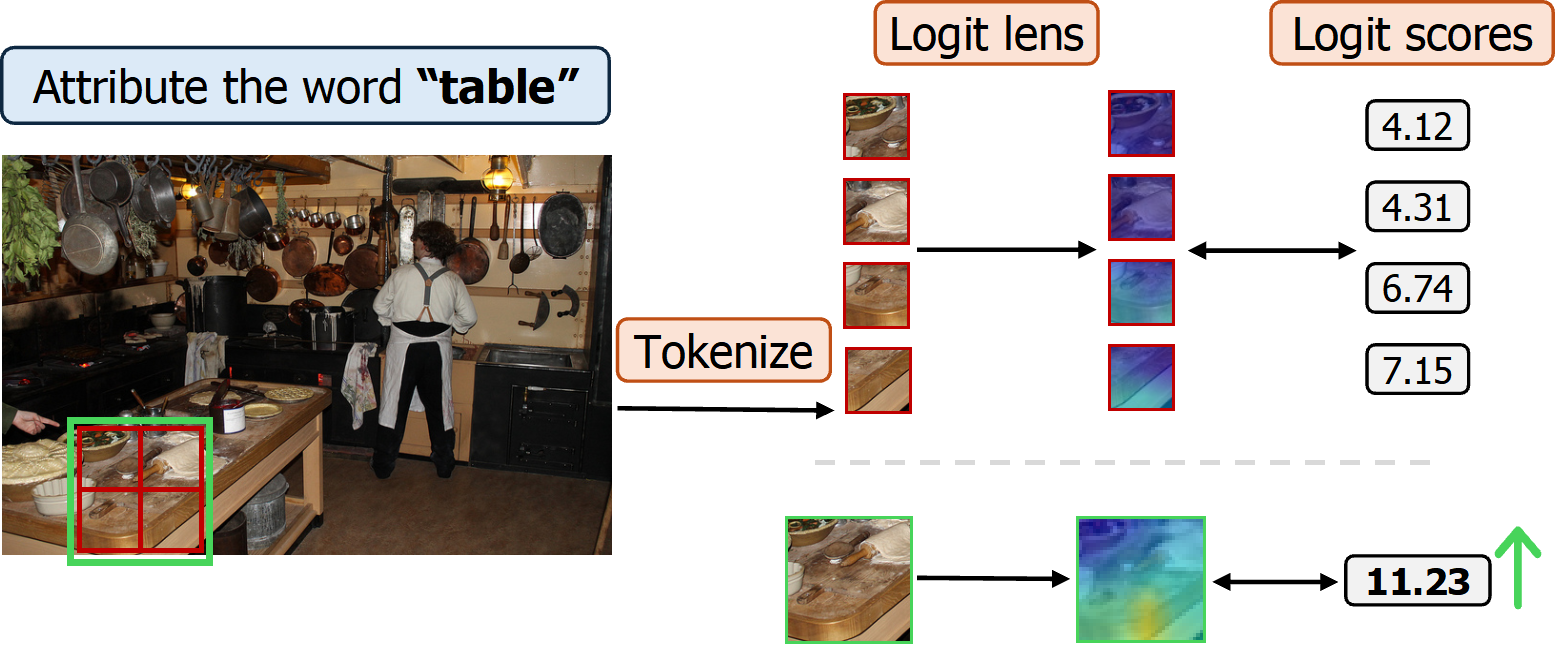}
        \label{fig:motivation_er}}
    \hfil
    \subfloat[Preceding-token context interferes with target-token attribution.]{
        \includegraphics[width=0.54\linewidth]{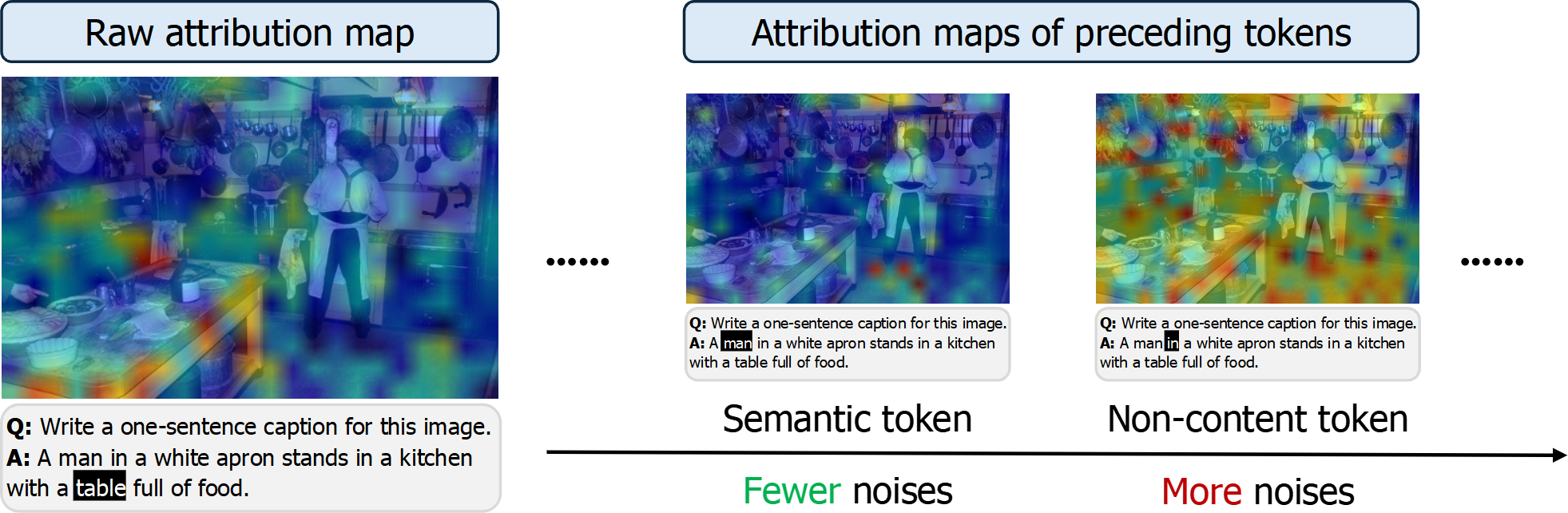}
        \label{fig:motivation_pcr}}
    \caption{Motivation. Logit-lens attribution decodes context-mixed visual tokens independently, making single-view attribution sensitive to token-to-region assignment. In addition, preceding tokens can interfere with the attribution of later target tokens.}
    \label{fig:motivation}
\end{figure*}

We propose \textbf{Evidence Recomposition and Predictive Context Residualization} (\method), which refines the logit-lens attribution interface to improve the stability of token-level visual evidence. The first component, \textbf{Evidence Recomposition} (\er), constructs multiple evidence views with different token-to-region assignments and applies the same target-token readout to each view. It aggregates recurring target evidence to reduce dependence on a single visual readout grid. The second component, \textbf{Predictive Context Residualization} (\pcr), uses RBO-based rank relevance to weight preceding-token attribution maps, forms a context map, and subtracts its fitted component from the ER map to mitigate preceding-token context interference.

We evaluate \method on LLaVA~\cite{liu2024visual}, Qwen2-VL~\cite{wang2024qwen2}, and InternVL~\cite{chen2024expanding,zhu2025internvl3,wang2025internvl35} families across COCO Caption~\cite{chen2015microsoft}, GranDf~\cite{rasheed2024glamm}, and OpenPSG~\cite{zhou2024openpsg}. Under the TAM-compatible evaluation protocol, \method improves visual evidence for target tokens and mitigates preceding-token context interference, surpassing existing attribution baselines across all three datasets. In particular, it improves F1-IoU over TAM by 5.35 and 6.37 points on COCO Caption and GranDf, respectively.

The main contributions are:
\begin{itemize}
    \item We identify two attribution-interface instabilities in logit-lens MLLM attribution: single-grid readout fragmentation from context-mixed visual tokens and preceding-token context interference.
    \item We propose Evidence Recomposition, which aggregates evidence across multiple token-to-region assignments to reduce dependence on a single visual readout grid for the same generated target token.
    \item We introduce Predictive Context Residualization, an RBO-based context residualization procedure that estimates preceding-token context and subtracts its fitted component from the ER map to mitigate preceding-token context interference.
    \item We validate \method across LLaVA, Qwen2-VL, and InternVL families on three benchmarks, consistently outperforming existing attribution baselines.
\end{itemize}

\section{Related Work}
\label{sec:related_work}

\subsection{Multimodal Large Language Models}
MLLMs combine visual encoders with autoregressive language models for visual question answering~\cite{antol2015vqa}, image captioning~\cite{chen2015microsoft,hossain2019comprehensive}, and open-ended visual reasoning. Flamingo~\cite{alayrac2022flamingo} injects visual features through gated cross-attention, while LLaVA~\cite{liu2024visual} aligns visual features with an instruction-tuned language model through a projection module. Recent systems such as Qwen2-VL~\cite{wang2024qwen2} and InternVL~\cite{chen2024internvl,chen2024expanding} improve visual resolution handling, backbone scale, and alignment. Their autoregressive and open-vocabulary generation makes attribution more difficult than in closed-set visual classification.

\subsection{Visual Attribution}
Visual attribution has been extensively studied for convolutional and transformer-based vision models~\cite{chen2024less,chen2025interpreting}. CAM~\cite{zhou2016learning}, Grad-CAM~\cite{selvaraju2017grad}, Grad-CAM++~\cite{chattopadhay2018grad}, and Layer-CAM~\cite{jiang2021layercam} localize image regions with class-discriminative activations or gradients. Perturbation and surrogate methods such as LIME~\cite{ribeiro2016should} and SHAP~\cite{lundberg2017unified} estimate feature importance from prediction changes~\cite{chen2025less}, while relevance-propagation methods trace importance through attention blocks or residual paths~\cite{lapuschkin2019unmasking,abnar2020quantifying,chefer2021transformer,chefer2021generic,ali2022xai,achtibat2024attnlrp}. These methods are often designed for classification or encoder-style prediction~\cite{chen2026wherenot,chen2025did,jiao2025subset}; MLLM attribution explains generated tokens conditioned on both visual evidence and preceding text.

The evaluation of attribution maps is also an image-space problem~\cite{yang2026can}. Prior work emphasizes that explanations should be tied to a specified target function and tested beyond visual plausibility~\cite{sundararajan2017axiomatic,adebayo2018sanity,jacovi2020towards}. IoU-based plausibility metrics compare thresholded maps with object masks or noun-derived foreground thresholds, while deletion and insertion test whether high-scoring regions influence the target prediction under controlled perturbations~\cite{petsiuk2018rise,zeiler2014visualizing}. These protocols measure complementary properties: spatial agreement, suppression of low-relevance visual evidence, and target-score sensitivity.

\subsection{Attribution for Vision-Language and Multimodal Models}
Interpretability for vision-language models has evolved from model inspection tools to token-level attribution for MLLM outputs~\cite{chen2026whereattend}. VL-Interpret~\cite{aflalo2022vl} and LVLM-Interpret~\cite{ben2024lvlm} analyze how vision-language transformers use visual and textual inputs. LLaVA-CAM~\cite{zhang2024redundancy} associates image regions with LLaVA responses, PROJECTAWAY~\cite{jiang2024interpreting} studies hallucination-related representations, and TAM~\cite{li2025token} decodes visual-token hidden states with the language unembedding matrix while suppressing contextual interference. These methods motivate two practical axes for our work: reducing dependence on a single visual readout grid and controlling context interference in token-level attribution. \method addresses these axes by aggregating evidence across multiple views and residualizing an RBO-based context map.

Logit-lens analysis~\cite{nostalgebraist2020logitlens} projects hidden states into the vocabulary space with the final unembedding matrix, allowing visual-token states to be scored against generated words. Related probing work such as the tuned lens~\cite{belrose2023eliciting} studies hidden-state prediction interfaces with learned translators. For MLLMs, this interface directly links visual hidden states to open-vocabulary output tokens, but it is sensitive to how context-mixed hidden states are assigned back to image locations and to the preceding text context used for token prediction. Context analysis also shows that preceding tokens can strongly shape later predictions~\cite{zhou2024mystery,zaranis2024analyzing}.

\section{Method}
\label{sec:method}

\begin{figure*}[!t]
    \centering
    \includegraphics[width=\linewidth]{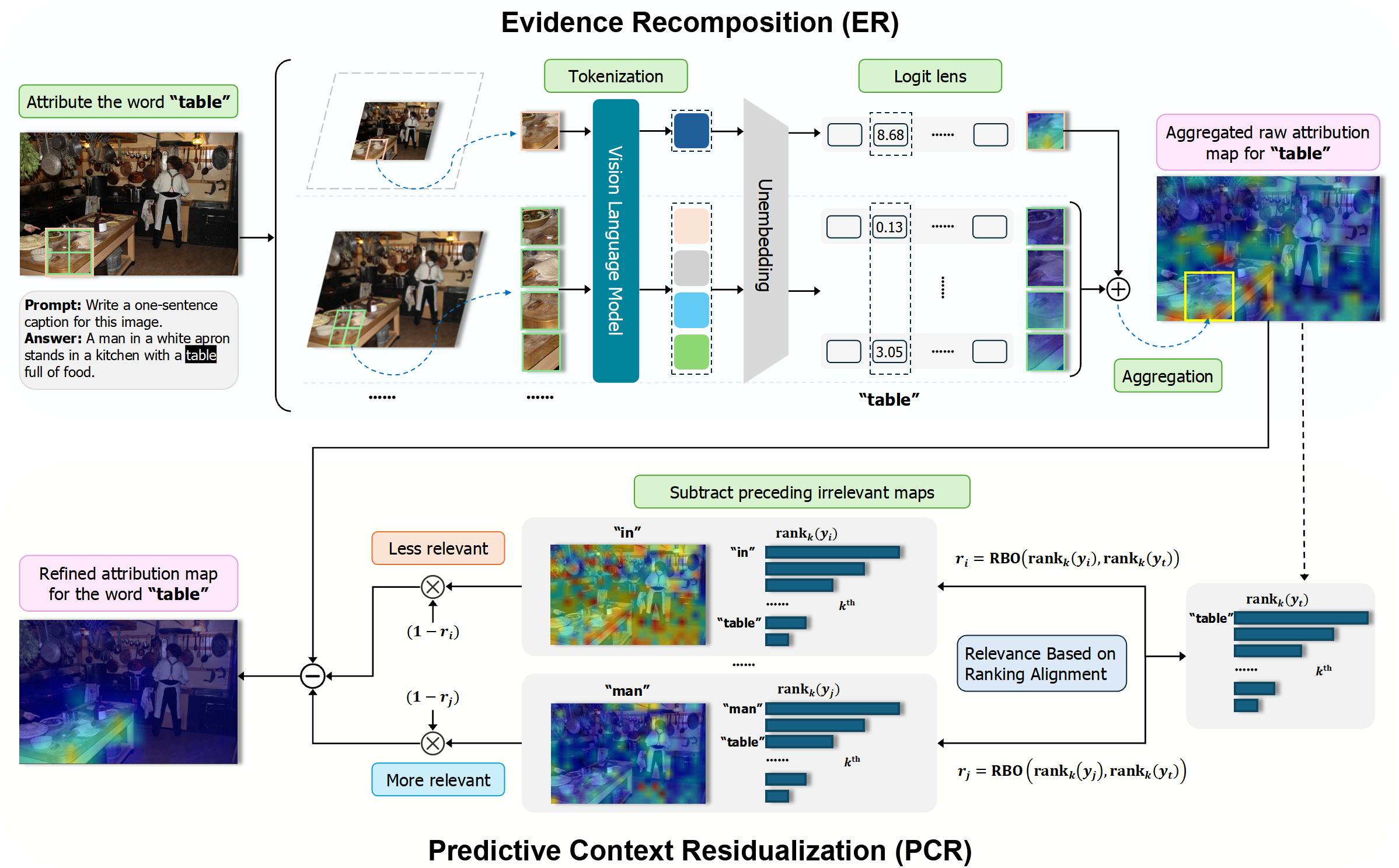}
    \caption{Conceptual overview of \method. Evidence Recomposition aggregates target support across multiple evidence views. Predictive Context Residualization estimates a preceding-token context map with RBO-based rank relevance and suppresses its interference in the attribution of the current target token.}
    \label{fig:framework}
\end{figure*}

\subsection{Preliminaries}
\label{subsec:prelim}

\textbf{MLLM generation.}
Given an input image $I$ and a sequence of preceding text tokens $T_{<t}=\{T_1,\ldots,T_{t-1}\}$, an MLLM predicts the next token $T_t$ with a multimodal transformer. The image is encoded into $N$ visual tokens $\{v_i\}_{i=1}^{N}$ and concatenated with text tokens. We denote by $\zv_i^l$ the hidden state of visual token $v_i$ at layer $l$, and by $\htxt_t^L$ the final-layer generation hidden state whose language-head projection emits $T_t$. With unembedding matrix $\WU$, the next-token distribution is
\begin{equation}
P(T_t \mid T_{<t}, I)=\softmax(\WU\htxt_t^L).
\label{eq:mllm}
\end{equation}
 
\textbf{Token-level logit-lens attribution.}
For a target token $T_t$ with vocabulary index $k_t$, logit-lens attribution projects each visual token hidden state into the vocabulary space and selects the target-token logit:
\begin{equation}
a_i^{l}(t)=\left[\WU \zv_i^l\right]_{k_t},
\label{eq:logit_lens}
\end{equation}
where $[\cdot]_{k_t}$ selects the target vocabulary coordinate. Arranging $\{a_i^{l}(t)\}_{i=1}^{N}$ according to the visual token grid gives a raw attribution map $\Araw_t$. In practice, we use the same layer choice as the base logit-lens attribution implementation unless otherwise specified.

\textbf{Readout target and response score.}
The vocabulary coordinate $k_t$ defines the readout target of the attribution map. We separately denote the response-level score for the same fixed prefix and target token as
\begin{equation}
s_t(I)=\log P(T_t\mid T_{<t}, I).
\label{eq:target_score}
\end{equation}
The anchor response fixes $T_{<t}$, $T_t$, and $k_t$; ER applies the same readout coordinate across transformed evidence views, and PCR refines the resulting recomposed map for that target.

\subsection{Two Attribution Instabilities}
\label{subsec:bottleneck}

Logit-lens attribution exposes two practical instabilities in its readout interface. First, a context-mixed visual-token hidden state is decoded independently and assigned back to a single grid cell. Evidence from a broader semantic region can therefore be tied to one visual token and appear as fragmented attribution.

\textbf{Single-grid readout fragmentation.}
Let $S_t\subseteq I$ denote the image support that provides visual evidence for a target token $T_t$. A token grid induces an assignment from image coordinates to visual tokens. If $S_t$ overlaps several token cells, the token-wise readout in Eq.~\eqref{eq:logit_lens} can distribute the same underlying evidence differently across token-to-region assignments. The resulting map can fragment target support. Our objective is to obtain a less grid-dependent evidence map for the same target token by recomposing attribution views produced under different assignments.

Second, the target token $T_t$ is predicted from the image and preceding text tokens $T_{<t}$. Under this autoregressive prediction context, the visual-token readout for $k_t$ can produce a map that shares spatial patterns with attribution maps of preceding tokens, including earlier words, punctuation, or connectors. We refer to this overlap as preceding-token context interference. Such overlap can introduce diffuse activation or activation unrelated to the current target. ER and PCR address these two attribution instabilities, respectively.

\begin{algorithm}[!t]
\caption{\method for Target Token Attribution}
\label{alg:ercr}
\begin{algorithmic}[1]
\STATE \textbf{Input:} Image $I$, generated target token $T_t$, evidence transforms $\{\phi_m\}_{m=1}^M$, layer $l$, top-$k$ size $k$
\STATE \textbf{Output:} Refined attribution map $\Atilde_t$
\FOR{$m=1$ to $M$}
    \STATE Construct evidence view $I^{(m)}=\phi_m(I)$
    \STATE Encode $I^{(m)}$ and extract visual hidden states $\{\zv_i^{l,(m)}\}_{i=1}^{N_m}$
    \STATE Compute $\Araw_t^{(m)}$ by Eq.~\eqref{eq:view_attr}
    \STATE Map $\Araw_t^{(m)}$ back to original image coordinates with $\warp_m$
\ENDFOR
\STATE Recompose $\Araw_t^{\mathrm{ER}}$ by Eq.~\eqref{eq:er}
\FOR{each preceding token $T_j$, $j<t$}
\STATE Compute rank relevance $r_j$ by Eq.~\eqref{eq:rank_relevance}
    \STATE Compute or retrieve ER attribution $\Araw_j^{\mathrm{ER}}$
\ENDFOR
\STATE Aggregate preceding-token context map $\Ahat_t$ by Eq.~\eqref{eq:residue}
\STATE Compute $\beta_t$ by Eq.~\eqref{eq:beta}
\STATE Return $\Atilde_t$ by Eq.~\eqref{eq:clean}
\end{algorithmic}
\end{algorithm}

\subsection{Evidence Recomposition}
\label{subsec:er}

ER reduces dependence on a single visual readout grid by recomposing multiple attribution maps for a fixed target token. It constructs evidence views through image transformations that change token-to-region assignments. Every view is read using the same target vocabulary coordinate fixed by the anchor response. The resulting maps expose spatial evidence under different visual-token partitions, and their consensus favors evidence that recurs across views.

\textbf{Evidence views.}
Let $\View=\{1,\ldots,M\}$ index visual evidence views. Each view corresponds to an image preprocessing transformation $\phi_m$ that changes how image regions are assigned to visual tokens for the same generated target token $T_t$. The concrete view family is specified in the experimental setup. The transformed image is
\begin{equation}
I^{(m)} = \phi_m(I).
\label{eq:evidence_view}
\end{equation}
For each evidence view, we encode $I^{(m)}$ and compute the logit-lens attribution map for the same target token $T_t$:
\begin{equation}
\Araw_t^{(m)}=\left\{ \left[\WU \zv_i^{l,(m)}\right]_{k_t} \right\}_{i=1}^{N_m}.
\label{eq:view_attr}
\end{equation}
The anchor image view provides the generated response, target token $T_t$, and vocabulary index $k_t$. Auxiliary transformed views expose visual-token hidden states under altered token-to-region assignments and provide view-specific readouts for the same anchor target.

\textbf{Recomposition.}
Before aggregation, each view-specific attribution $\Araw_t^{(m)}$ is represented on the original image grid by a view-dependent mapping operator $\warp_m(\cdot)$:
\begin{equation}
\bar{\Araw}_t^{(m)}=\warp_m\left(\Araw_t^{(m)}\right).
\label{eq:warp}
\end{equation}
The operator $\warp_m$ is defined by the geometry of $\phi_m$. The resulting maps are min-max normalized over valid image tokens to reduce scale differences across views. The recomposed attribution map is
\begin{equation}
\Araw_t^{\mathrm{ER}}=\sum_{m=1}^{M} w_m \bar{\Araw}_t^{(m)}, \quad w_m\geq0,\quad \sum_{m=1}^{M}w_m=1.
\label{eq:er}
\end{equation}
Equivalently, ER is the weighted least-squares consensus map
\begin{equation}
\Araw_t^{\mathrm{ER}}
=\argmin_{\mathbf{A}}
\sum_{m=1}^{M} w_m
\left\|\mathbf{A}-\bar{\Araw}_t^{(m)}\right\|_2^2 .
\label{eq:er_consensus}
\end{equation}
Let $\bar{\mathbf{a}}_t^{(m)}=\operatorname{vec}(\bar{\Araw}_t^{(m)})$ and $\mathbf{a}_t^{\mathrm{ER}}=\sum_m w_m\bar{\mathbf{a}}_t^{(m)}$. For any candidate vector $\mathbf{A}$, the exact weighted variance decomposition is
\begin{equation}
\sum_m w_m\|\mathbf{A}-\bar{\mathbf{a}}_t^{(m)}\|_2^2
=\|\mathbf{A}-\mathbf{a}_t^{\mathrm{ER}}\|_2^2
+\sum_m w_m\|\bar{\mathbf{a}}_t^{(m)}-\mathbf{a}_t^{\mathrm{ER}}\|_2^2 .
\label{eq:er_variance_decomp}
\end{equation}
Unless stated otherwise, we use uniform weights $w_m=1/M$. For view-specific maps scored with the same target vocabulary index, ER uniquely minimizes finite-view disagreement. Rather than selecting a single view, Eq.~\eqref{eq:er_variance_decomp} favors evidence that recurs across views and reduces the influence of isolated view-specific peaks. Shared non-target patterns can remain after recomposition, which motivates residualization against preceding-token context in PCR.

\subsection{Predictive Context Residualization}
\label{subsec:pcr}

PCR estimates and removes the preceding-token context component from the attribution of the current target. In our implementation, the preceding-token set is formed from textual positions before the explained target in the decoded prompt-answer token stream, excluding image tokens. First, we use RBO-based rank relevance as an operational proxy for how closely each preceding-token prediction distribution relates to the current target prediction distribution. Preceding tokens with lower rank relevance are assigned larger context weights, because their maps are less likely to share target evidence and more likely to capture context-specific residue. These weights determine how the preceding-token maps form the context map. After this map is constructed, a coefficient for the current target is fitted to estimate how strongly the context map is represented in the current ER map. PCR then subtracts the fitted component.

For each preceding token $T_j$, $j<t$, we compute its vocabulary distribution from the final-layer text hidden state:
\begin{equation}
\mathbf{y}_j=\softmax(\WU\htxt_j^L).
\label{eq:text_logits}
\end{equation}
Let $\rank_k(\mathbf{y}_j)$ be the ordered list of the top-$k$ vocabulary indices. We define the rank relevance $r_j$ between the prediction distributions at positions $j$ and $t$ by Rank-Biased Overlap (RBO) over their top-$k$ vocabulary lists~\cite{webber2010similarity}:
\begin{equation}
r_j=\RBO\left(\rank_k(\mathbf{y}_j),\rank_k(\mathbf{y}_t)\right).
\label{eq:rank_relevance}
\end{equation}
For two ranked lists $R_1$ and $R_2$, RBO is
\begin{equation}
\RBO(R_1,R_2)=(1-p)\sum_{d=1}^{\infty}p^{d-1}
\frac{|R_1(1{:}d)\cap R_2(1{:}d)|}{d},
\label{eq:rbo}
\end{equation}
where $p\in[0,1)$ controls the top-rank bias. This score provides the RBO-based rank relevance used to construct the preceding-token context map.
In implementation, the infinite summation is evaluated on the top-$k$ lists used by $\rank_k(\cdot)$, with the standard finite-depth RBO approximation. This keeps the comparison focused on high-probability vocabulary candidates and avoids making the score sensitive to the long tail of the vocabulary distribution.

We convert lower rank relevance into normalized context weights
\begin{equation}
\alpha_{jt}=
\frac{1-r_j}
{\sum_{\ell<t}(1-r_\ell)+\epsilon},
\quad j<t .
\label{eq:context_weight}
\end{equation}
We compute an ER attribution map $\Araw_j^{\mathrm{ER}}$ for each preceding token and aggregate a preceding-token context map by
\begin{equation}
\Ahat_t =
\sum_{j<t}\alpha_{jt}\Araw_j^{\mathrm{ER}} .
\label{eq:residue}
\end{equation}
After constructing $\Ahat_t$, we fit the scalar
\begin{equation}
\beta_t=\argmin_{\beta}\left\|\Araw_t^{\mathrm{ER}}-\beta\Ahat_t\right\|_2^2
\label{eq:beta}
\end{equation}
to estimate how strongly the context map is represented in the current ER map. We use the closed-form least-squares solution
\begin{equation}
\beta_t =
\frac{\left\langle \Araw_t^{\mathrm{ER}}, \Ahat_t \right\rangle}
{\left\langle \Ahat_t, \Ahat_t \right\rangle+\epsilon},
\label{eq:beta_closed}
\end{equation}
as an unconstrained scalar least-squares coefficient. PCR then forms the pre-clipping residual
\begin{equation}
\mathbf{R}_t=\Araw_t^{\mathrm{ER}}-\beta_t\Ahat_t
\label{eq:pcr_residual}
\end{equation}
with respect to the estimated preceding-token context map. When $\epsilon$ is negligible and $\Ahat_t$ is nonzero, Eq.~\eqref{eq:beta_closed} gives the normal-equation condition $\langle \mathbf{R}_t,\Ahat_t\rangle=0$ for this pre-clipping residual. The residual evidence map is then positive-clipped and passed to the rank Gaussian filter:
\begin{equation}
\Atilde_t=\mathcal{G}\left(\pospart{\mathbf{R}_t}\right),
\label{eq:clean}
\end{equation}
where $\pospart{\cdot}$ keeps positive evidence, and $\mathcal{G}$ is a rank Gaussian filter that sorts local scores and applies Gaussian weights over their ranks to reduce isolated local peaks. The local projection analysis below is stated for the pre-filter residual $\mathbf{R}_t$.

\subsection{Analysis of ER and PCR}
\label{subsec:er_pcr_analysis}

The preceding formulas give two properties: ER is the weighted consensus in Eqs.~\eqref{eq:er_consensus}--\eqref{eq:er_variance_decomp}, and PCR is least-squares residualization with respect to the preceding-token context map. Let $\mathbf{a}_t=\operatorname{vec}(\Araw_t^{\mathrm{ER}})$ and $\mathbf{c}_t=\operatorname{vec}(\Ahat_t)$. Ignoring the small numerical stabilizer for notation, Eq.~\eqref{eq:beta_closed} subtracts the projection of $\mathbf{a}_t$ onto the estimated context-map direction:
\begin{equation}
\mathbf{P}_{\mathbf{c}}=
\frac{\mathbf{c}_t\mathbf{c}_t^{\top}}{\|\mathbf{c}_t\|_2^2},
\qquad
\mathbf{R}_t=(\mathbf{I}-\mathbf{P}_{\mathbf{c}})\mathbf{a}_t .
\label{eq:projection_residual}
\end{equation}
This projection form is the vectorized version of Eq.~\eqref{eq:pcr_residual}. It gives the closest point to $\mathbf{a}_t$ in $\mathbf{c}_t^{\perp}$ and satisfies $\langle \mathbf{R}_t,\mathbf{c}_t\rangle=0$ when $\epsilon=0$ and $\mathbf{c}_t\neq\mathbf{0}$. Eq.~\eqref{eq:beta_closed} is the finite-$\epsilon$ implementation used in PCR. We use the control-variate link as a least-squares residualization analogy~\cite{owen2013monte}.

\textbf{Projection stability.}
Suppose locally that $\mathbf{a}_t=\boldsymbol{\theta}+\gamma\mathbf{c}_t+\mathbf{e}$, where $\boldsymbol{\theta}$ is the stable target component and $\mathbf{e}$ collects finite-view and higher-order residuals. If $\epsilon=0$, $\mathbf{c}_t\neq\mathbf{0}$, and $\boldsymbol{\theta}\perp\mathbf{c}_t$, then
\begin{equation}
\mathbf{R}_t-\boldsymbol{\theta}
=(\mathbf{I}-\mathbf{P}_{\mathbf{c}})\mathbf{e},
\quad
\|\mathbf{R}_t-\boldsymbol{\theta}\|_2\leq\|\mathbf{e}\|_2 .
\label{eq:projection_stability}
\end{equation}
Thus the projection subtracts the fitted context-map component without amplifying the residual norm when the target component is orthogonal to $\mathbf{c}_t$; otherwise alignment between target and context appears as the bias $-\mathbf{P}_{\mathbf{c}}\boldsymbol{\theta}$. This local projection view applies to the pre-clipping residual $\mathbf{R}_t$ and relies on local linearity, an estimated one-dimensional context-map direction, and bounded residual terms. The positive clipping and rank filter in Eq.~\eqref{eq:clean} are applied after this residualization step.

\begin{table*}[!t]
\centering
\footnotesize
\setlength{\tabcolsep}{3.0pt}
\renewcommand{\arraystretch}{1.0}
\caption{Comparison with attribution baselines on Qwen2-VL-2B under the TAM-compatible scoring protocol.}
\label{tab:sota}
\begin{tabularx}{\textwidth}{lc|YYY|YYY|YYY}
\toprule
\multirow{2}{*}{Method} & \multirow{2}{*}{Type} &
\multicolumn{3}{c|}{COCO Caption} &
\multicolumn{3}{c|}{GranDf} &
\multicolumn{3}{c}{OpenPSG} \\
 & & Obj-IoU & Func-IoU & F1-IoU & Obj-IoU & Func-IoU & F1-IoU & Obj-IoU & Func-IoU & F1-IoU \\
\midrule
Grad-CAM++~\cite{chattopadhay2018grad} & \multirow{3}{*}{Gradient} & 19.52 & 62.83 & 29.78 & 17.30 & 73.42 & 28.01 & 22.21 & 59.95 & 32.41 \\
Grad-Rollout~\cite{abnar2020quantifying} & & 1.27 & 99.51 & 2.51 & 1.40 & 99.61 & 2.77 & 1.57 & 99.58 & 3.09 \\
Layer-CAM~\cite{jiang2021layercam} & & 11.43 & 84.88 & 20.15 & 13.11 & 82.09 & 22.62 & 14.12 & 85.29 & 24.22 \\
\midrule
Attention~\cite{vaswani2017attention} & \multirow{3}{*}{Attention} & 8.20 & 92.87 & 15.07 & 9.60 & 93.56 & 17.42 & 10.58 & 94.28 & 19.03 \\
Attention-Rollout~\cite{abnar2020quantifying} & & 5.74 & 96.50 & 10.83 & 7.21 & 96.65 & 13.42 & 7.94 & 97.04 & 14.68 \\
LVLM-Interpret~\cite{ben2024lvlm} & & 9.99 & 82.90 & 17.82 & 10.87 & 89.65 & 19.40 & 15.80 & 83.87 & 26.59 \\
\midrule
VL-InterpreT~\cite{aflalo2022vl} & Repr. & 16.31 & 39.53 & 23.10 & 17.05 & 46.83 & 25.00 & 20.12 & 41.87 & 27.18 \\
\midrule
CP-LRP~\cite{ali2022xai} & \multirow{2}{*}{Hybrid} & 9.90 & 53.97 & 16.73 & 12.61 & 53.24 & 20.39 & 13.30 & 53.36 & 21.30 \\
Attn-LRP~\cite{achtibat2024attnlrp} & & 9.92 & 52.41 & 16.69 & 12.15 & 52.19 & 19.72 & 12.78 & 52.26 & 20.54 \\
\midrule
CAM~\cite{zhou2016learning} & \multirow{5}{*}{Logit} & 21.23 & 51.93 & 30.14 & 17.85 & 62.15 & 27.74 & 22.93 & 48.57 & 31.15 \\
LLaVA-CAM~\cite{zhang2024redundancy} & & 24.78 & 40.14 & 30.64 & 21.50 & 29.89 & 25.01 & 27.02 & 22.79 & 24.73 \\
Archi.-Surgery~\cite{li2025closer} & & 15.69 & 63.82 & 25.19 & 16.59 & 62.28 & 26.20 & 19.83 & 58.77 & 29.65 \\
TAM~\cite{li2025token} & & 27.37 & 68.44 & \underline{39.10} & 18.65 & 88.97 & \underline{30.83} & 26.26 & 92.99 & \underline{40.95} \\
\rowcolor{lightgray}\textbf{\method} & & 29.35 & 91.57 & \textbf{44.45} & 23.32 & 91.85 & \textbf{37.20} & 29.20 & 94.69 & \textbf{44.64} \\
\bottomrule
\end{tabularx}

\end{table*}

\subsection{Algorithm and Complexity}
\label{subsec:algorithm}

Algorithm~\ref{alg:ercr} summarizes the complete pipeline. Let $C_{\mathrm{probe}}$ be the cost of computing one visual attribution view, $M$ be the number of evidence views, $t$ be the number of preceding text tokens for the current target, and $k$ be the ranking depth. ER costs $O(M C_{\mathrm{probe}})$ per attributed token before reuse. For a single target token, PCR adds top-$k$ rank relevance comparisons with cost $O(tk)$ and uses preceding-token attribution maps for residue estimation. In sequence-level attribution, previously computed maps can be reused. In practice, the ranking overhead is small relative to the visual evidence probes. The number of views directly controls the main accuracy-efficiency trade-off.

\section{Experiments}
\label{sec:experiments}

This section first specifies the evaluation protocol, then compares \method with attribution baselines, tests generalization across architectures and scales, analyzes ER/PCR components and evidence-view design, evaluates fixed-prefix perturbation faithfulness, and reports runtime, memory, and counted FLOPs.

\subsection{Experimental Setup}
\label{subsec:setup}

\textbf{Models and datasets.}
We evaluate \method on LLaVA-1.5~\cite{liu2024visual}, Qwen2-VL~\cite{wang2024qwen2}, and InternVL2.5~\cite{chen2024expanding}, covering model scales from 2B to 13B parameters. We also include InternVL3~\cite{zhu2025internvl3} and InternVL3.5~\cite{wang2025internvl35} at the 2B scale for cross-generation validation. The primary benchmark is COCO Caption~\cite{chen2015microsoft} with COCO 2014~\cite{lin2014microsoft} minival segmentation annotations. We further evaluate on GranDf~\cite{rasheed2024glamm} with 1K images and OpenPSG~\cite{zhou2024openpsg} with 3,176 validation images. COCO Caption and GranDf provide human masks, while OpenPSG masks are derived from integrated annotations.

\textbf{Implementation details.}
All methods use dataset-specific prompts following the TAM-compatible benchmark protocol~\cite{li2025token}; for example, COCO Caption uses ``Write a one-sentence caption for this image:''. Attribution is computed for generated tokens under the same model preprocessing. Unless otherwise specified, ER uses scale factors $[0.5,0.75,1.0]$ with mean aggregation. PCR uses top-$k=50$ vocabulary lists for rank relevance, RBO decay $p=0.8$, $\epsilon=10^{-8}$, and a $3{\times}3$ rank Gaussian filter after positive clipping. The logit-lens layer follows the corresponding baseline setting.

\begin{table*}[!t]
\centering
\footnotesize
\setlength{\tabcolsep}{5pt}
\renewcommand{\arraystretch}{1.0}
\caption{Generalization across MLLM architectures, model scales, and InternVL generations.}
\label{tab:scalability}
\begin{tabularx}{\textwidth}{lc|YYY|YYY|YYY}
\toprule
\multirow{2}{*}{Method} & \multirow{2}{*}{MLLM} &
\multicolumn{3}{c|}{COCO Caption} &
\multicolumn{3}{c|}{GranDf} &
\multicolumn{3}{c}{OpenPSG} \\
 & & Obj-IoU & Func-IoU & F1-IoU & Obj-IoU & Func-IoU & F1-IoU & Obj-IoU & Func-IoU & F1-IoU \\
\midrule
CAM & \multirow{3}{*}{LLaVA1.5-7B} & 23.17 & 43.16 & 30.16 & 20.07 & 47.48 & 28.21 & 25.11 & 51.55 & 33.77 \\
TAM & & 27.65 & 61.43 & 38.13 & 20.71 & 59.15 & 30.68 & 28.57 & 61.06 & 38.93 \\
\rowcolor{lightgray}\method & & 30.62 & 87.32 & 45.34 & 24.79 & 85.18 & 38.40 & 32.03 & 86.80 & 46.79 \\
\midrule
CAM & \multirow{3}{*}{LLaVA1.5-13B} & 24.82 & 51.18 & 33.43 & 21.34 & 43.99 & 28.74 & 26.65 & 48.45 & 34.39 \\
TAM & & 29.12 & 58.50 & 38.88 & 22.10 & 51.02 & 30.84 & 30.88 & 59.96 & 40.76 \\
\rowcolor{lightgray}\method & & 31.76 & 97.18 & 47.87 & 26.08 & 95.58 & 40.98 & 32.57 & 97.32 & 48.80 \\
\midrule
CAM & \multirow{3}{*}{InternVL2.5-2B} & 15.94 & 45.62 & 23.63 & 18.28 & 37.64 & 24.61 & 19.76 & 46.42 & 27.72 \\
TAM & & 21.38 & 65.10 & 32.19 & 20.48 & 85.93 & 33.08 & 23.00 & 86.86 & 36.36 \\
\rowcolor{lightgray}\method & & 30.61 & 76.48 & 43.72 & 24.54 & 88.93 & 38.47 & 31.50 & 91.03 & 46.81 \\
\midrule
CAM & \multirow{3}{*}{InternVL2.5-4B} & 18.23 & 40.95 & 25.23 & 20.91 & 44.52 & 28.46 & 21.28 & 34.70 & 26.38 \\
TAM & & 21.76 & 63.12 & 32.36 & 22.53 & 89.71 & 36.02 & 23.49 & 89.75 & 37.23 \\
\rowcolor{lightgray}\method & & 31.80 & 82.73 & 45.94 & 27.73 & 94.34 & 42.86 & 33.52 & 94.09 & 49.43 \\
\midrule
CAM & \multirow{3}{*}{InternVL2.5-8B} & 14.59 & 64.41 & 23.80 & 18.04 & 57.42 & 27.45 & 18.46 & 62.21 & 28.47 \\
TAM & & 19.98 & 66.53 & 30.73 & 21.56 & 85.95 & 34.47 & 21.73 & 88.74 & 34.91 \\
\rowcolor{lightgray}\method & & 32.16 & 73.20 & 44.69 & 27.00 & 86.72 & 41.18 & 33.97 & 91.11 & 49.49 \\
\midrule
CAM & \multirow{3}{*}{InternVL3-2B} & 12.55 & 88.20 & 21.97 & 16.24 & 33.07 & 21.79 & 17.67 & 35.70 & 23.64 \\
TAM & & 23.04 & 65.18 & 34.05 & 21.27 & 86.50 & 34.14 & 24.25 & 92.91 & 38.46 \\
\rowcolor{lightgray}\method & & 29.57 & 84.92 & 43.86 & 26.44 & 92.29 & 41.10 & 30.78 & 92.43 & 46.18 \\
\midrule
CAM & \multirow{3}{*}{InternVL3.5-2B} & 14.94 & 19.44 & 16.90 & 15.44 & 25.08 & 19.12 & 17.96 & 27.84 & 21.83 \\
TAM & & 29.72 & 67.83 & 41.33 & 23.21 & 86.46 & 36.60 & 29.48 & 90.39 & 44.46 \\
\rowcolor{lightgray}\method & & 35.43 & 93.66 & 51.41 & 27.91 & 94.35 & 43.08 & 35.71 & 95.28 & 51.95 \\
\midrule
CAM & \multirow{3}{*}{Qwen2-VL-2B} & 21.23 & 51.93 & 30.14 & 17.85 & 62.15 & 27.74 & 22.93 & 48.57 & 31.15 \\
TAM & & 27.37 & 68.44 & 39.10 & 18.65 & 88.97 & 30.83 & 26.26 & 92.99 & 40.95 \\
\rowcolor{lightgray}\method & & 29.35 & 91.57 & 44.45 & 23.32 & 91.85 & 37.20 & 29.20 & 94.69 & 44.64 \\
\midrule
CAM & \multirow{3}{*}{Qwen2-VL-7B} & 22.51 & 42.44 & 29.42 & 18.60 & 68.03 & 29.21 & 23.41 & 42.94 & 30.30 \\
TAM & & 28.13 & 71.85 & 40.43 & 19.88 & 90.57 & 32.61 & 26.94 & 89.88 & 41.45 \\
\rowcolor{lightgray}\method & & 29.86 & 94.77 & 45.41 & 23.53 & 90.59 & 37.35 & 29.01 & 94.33 & 44.37 \\
\bottomrule
\end{tabularx}

\end{table*}

\textbf{Baselines.}
All baselines explain the same generated tokens under the same prompts and image preprocessing. We compare with gradient-based methods, including Grad-CAM++, Grad-Rollout, and Layer-CAM; attention-based methods, including raw attention, Attention-Rollout, and LVLM-Interpret; representation and hybrid methods, including VL-InterpreT, CP-LRP, and Attn-LRP; and logit-based methods, including CAM, LLaVA-CAM, Architecture Surgery, and TAM. \method belongs to the logit-lens family and keeps the same attribution interface while adding ER and PCR.

\textbf{Metrics.}
We report Obj-IoU, Func-IoU, and F1-IoU following the TAM-compatible IoU-based protocol~\cite{li2025token}. The protocol evaluates two complementary requirements: object words should localize the corresponding image regions, while function words and other tokens with low visual relevance should avoid high-confidence image evidence. Obj-IoU measures the first requirement by comparing binarized attribution maps with object masks. For object-associated tokens, let $\mathcal{B}(\Araw_i)$ be the binarized attribution map and $\mathbf{G}_i$ be the corresponding object mask. Obj-IoU is
\begin{equation}
\mathrm{Obj\text{-}IoU}=
\frac{1}{o}\sum_{i=1}^{o}
\frac{\left|\mathcal{B}(\Araw_i)\cap \mathbf{G}_i\right|}
{\left|\mathcal{B}(\Araw_i)\cup \mathbf{G}_i\right|}.
\label{eq:obj_iou}
\end{equation}
Func-IoU measures the second requirement for tokens treated as having low visual relevance by the protocol, where higher values indicate that fewer pixels exceed a noun-derived foreground threshold. With threshold $b_i$ defined from noun-token Otsu thresholds in the same image and image mask $\mathbf{J}$,
\begin{equation}
\mathrm{Func\text{-}IoU}=
\frac{1}{u}\sum_{i=1}^{u}
\frac{\left|(\Araw_i<b_i)\cap \mathbf{J}\right|}
{\left|(\Araw_i<b_i)\cup \mathbf{J}\right|}.
\label{eq:func_iou}
\end{equation}
Following the same protocol, we use F1-IoU as the overall plausibility score because it penalizes methods that improve only one side of the evaluation. A method with highly sparse maps can obtain high Func-IoU but poor visual evidence for object words, while a diffuse object-focused map can obtain better Obj-IoU but poor suppression of low-relevance visual evidence. The harmonic mean emphasizes methods that satisfy both requirements:
\begin{equation}
\text{F1-IoU}=
\frac{2\cdot\text{Obj-IoU}\cdot\text{Func-IoU}}
{\text{Obj-IoU}+\text{Func-IoU}}.
\label{eq:f1_exp}
\end{equation}
We use F1-IoU as the main score and report Obj-IoU and Func-IoU separately to assess object-mask agreement and suppression of low-relevance visual evidence. Attribution-map binarization, noun-derived foreground thresholds, and token-to-mask matching follow the TAM-compatible protocol. Unless otherwise stated, reported IoU-based metrics are computed on the full evaluation split of each dataset; deterministic subword variants and perturbation-based faithfulness evaluation use the settings stated in their captions.

\subsection{Comparison with Attribution Baselines}
\label{subsec:sota}

Table~\ref{tab:sota} compares \method with gradient-, attention-, representation-, hybrid-, and logit-based attribution methods on Qwen2-VL-2B. On the main F1-IoU score, \method improves over TAM by 5.35, 6.37, and 3.69 points on COCO Caption, GranDf, and OpenPSG, respectively. Both sides of the balanced metric improve: Obj-IoU increases on all three datasets, and Func-IoU rises most clearly on COCO from 68.44 to 91.57. The baseline families expose a two-sided failure mode: sparse attention maps may reduce interference but miss target evidence, whereas gradient maps recover more target evidence but retain more interference. Within the logit-lens interface, \method addresses these limitations by aggregating evidence across multiple views and mitigating preceding-token context interference.

\subsection{Generalization Across Architectures and Scales}
\label{subsec:scalability}

Table~\ref{tab:scalability} tests whether the gains persist across model families, scales, preprocessing pipelines, and InternVL generations. \method obtains the highest F1-IoU for every listed model and dataset. The improvement is especially clear on InternVL2.5, where COCO Obj-IoU increases from 21.38 to 30.61 on the 2B model and remains above 31 on the 4B and 8B models.

At a fixed 2B scale, the cross-generation InternVL results show that the effect persists across model releases. On InternVL3-2B, \method improves F1-IoU over TAM by 9.81, 6.96, and 7.72 points on COCO Caption, GranDf, and OpenPSG; on InternVL3.5-2B, the gains are 10.08, 6.48, and 7.49 points. These results indicate consistent gains across the tested backbones, model scales, and preprocessing pipelines.

\subsection{Ablation Studies}
\label{subsec:ablation}

\begin{table*}[!t]
\centering
\footnotesize
\setlength{\tabcolsep}{4.6pt}
\renewcommand{\arraystretch}{1.08}
\caption{Component ablation of ER and PCR on COCO Caption.}
\label{tab:ablation}
\begin{tabularx}{\textwidth}{YY|YYY|YYY|YYY}
\toprule
\multicolumn{2}{c|}{Evidence Recomposition} &
\multicolumn{3}{c|}{Predictive Context Residualization} &
\multicolumn{3}{c|}{Qwen2-VL-2B} &
\multicolumn{3}{c}{LLaVA1.5-7B} \\
\cmidrule(lr){1-2}\cmidrule(lr){3-5}
Mean & Max & PCR & TAM-style & Mean sub. &
Obj-IoU & Func-IoU & F1-IoU & Obj-IoU & Func-IoU & F1-IoU \\
\midrule
 & & & & & 24.82 & 43.34 & 31.57 & 25.41 & 39.74 & 31.00 \\
 & & & & \checkmark & 27.84 & 49.85 & 35.72 & 27.65 & 61.43 & 38.13 \\
 & & & \checkmark & & 27.37 & 68.44 & 39.10 & 27.81 & 66.22 & 39.17 \\
\rowcolor{lightgray} & & \checkmark & & & 27.07 & 91.43 & 41.78 & 28.04 & 85.15 & 42.19 \\
\midrule
& \checkmark & & & & 24.32 & 63.77 & 35.21 & 27.49 & 52.53 & 36.09 \\
\rowcolor{lightgray}\checkmark & & & & & 25.81 & 63.21 & 36.65 & 27.59 & 48.88 & 35.27 \\
\midrule
& \checkmark & \checkmark & & & 27.77 & 91.64 & 42.63 & 30.12 & 85.32 & 44.52 \\
\rowcolor{lightgray}\checkmark & & \checkmark & & & 29.35 & 91.57 & 44.45 & 30.62 & 87.32 & 45.34 \\
\bottomrule
\end{tabularx}

\end{table*}

\begin{figure*}[!t]
    \centering
    \subfloat[Qwen2-VL-2B]{
        \includegraphics[width=0.48\textwidth]{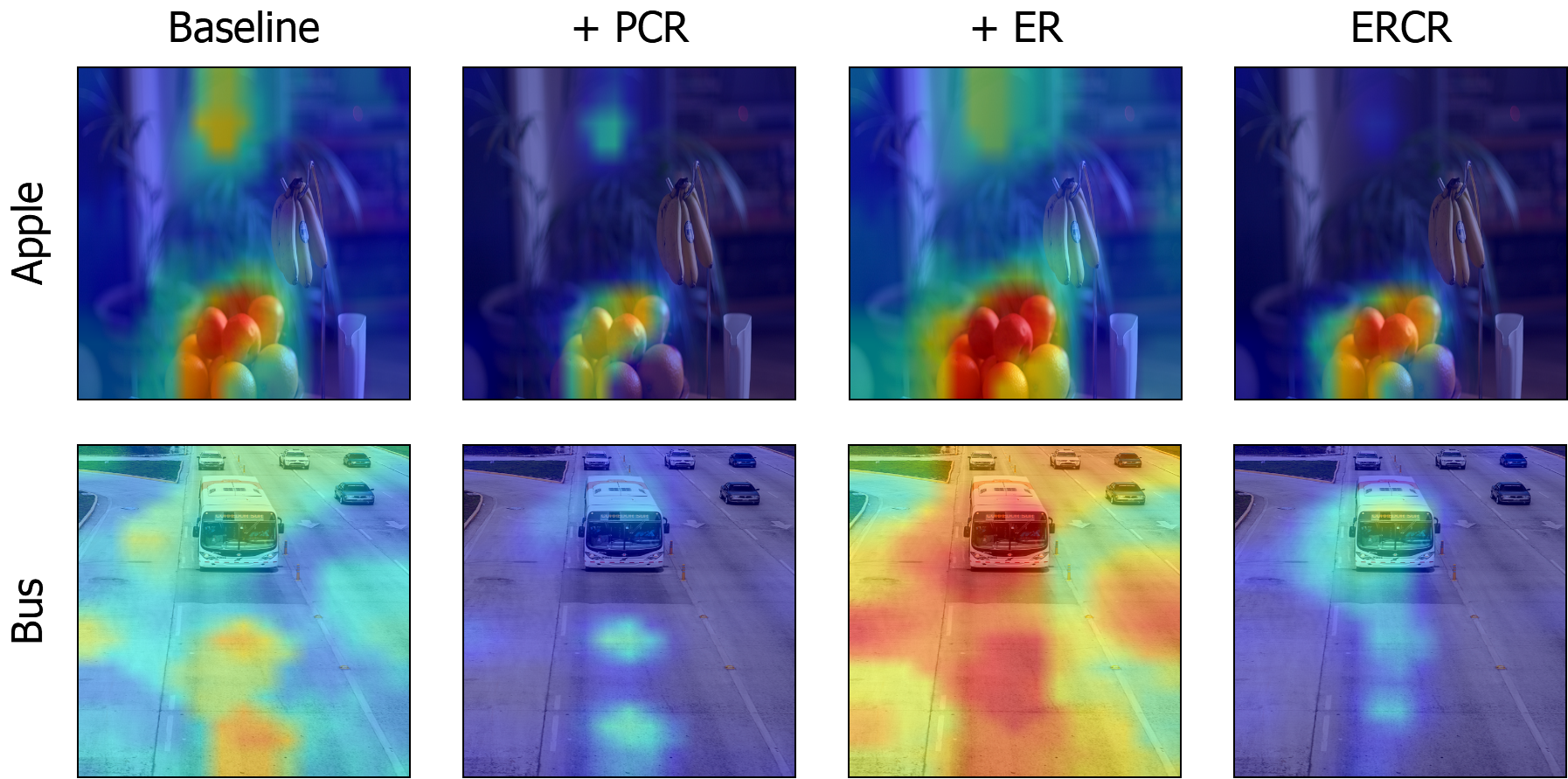}
    }\hfill
    \subfloat[InternVL2.5-2B]{
        \includegraphics[width=0.48\textwidth]{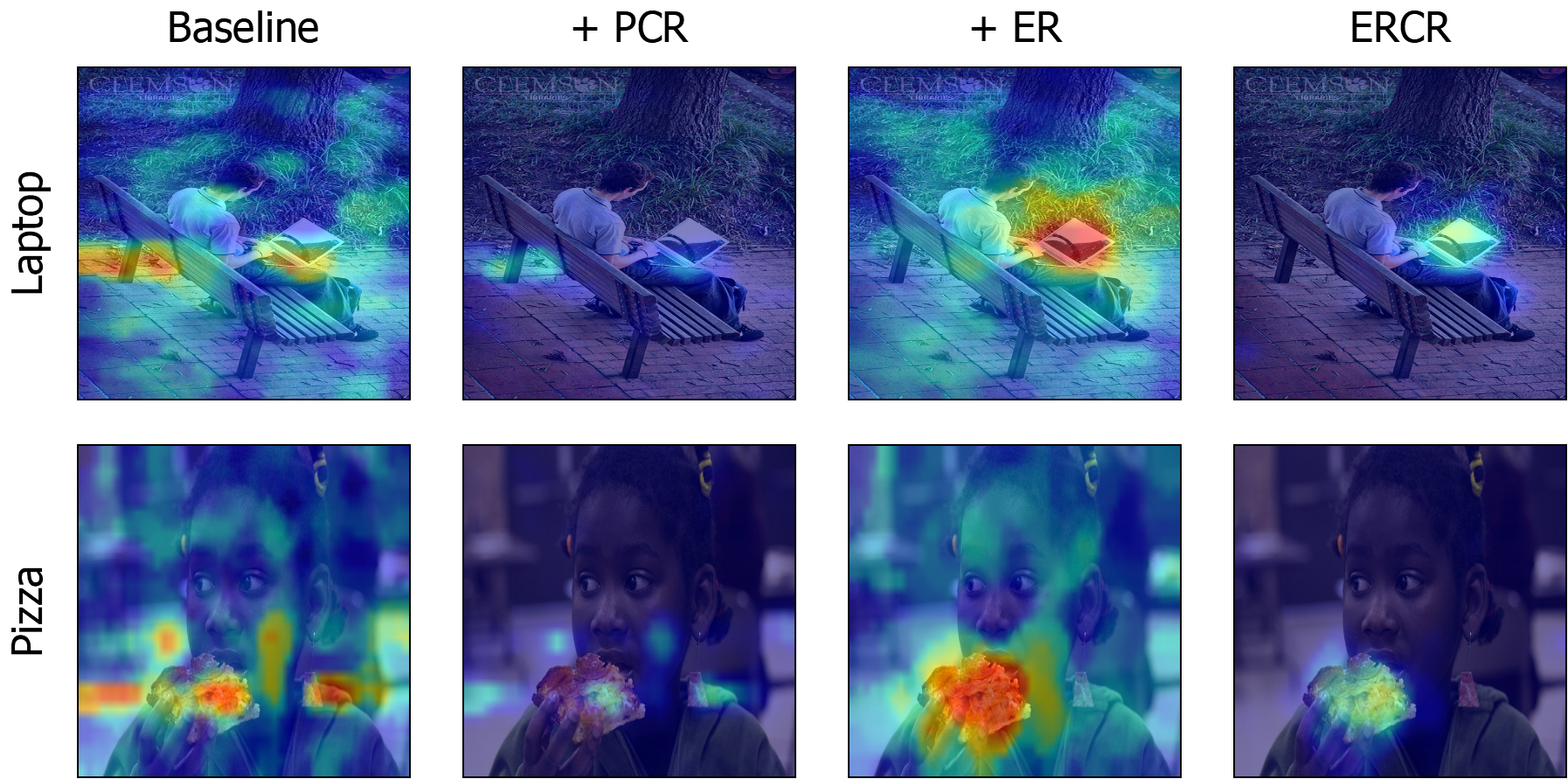}
    }
    \caption{Qualitative component ablation of ER and PCR. Columns compare Baseline (without ER or PCR), PCR only, ER only, and \method. ER aggregates evidence across views, while PCR mitigates context interference.}
    \label{fig:ablation_visual}
\end{figure*}

\begin{table}[!t]
\centering
\footnotesize
\setlength{\tabcolsep}{2.6pt}
\renewcommand{\arraystretch}{1.08}
\caption{Deterministic subword evaluation on COCO Caption.}
\label{tab:oracle_free_subword}
\begin{tabularx}{\columnwidth}{l>{\raggedright\arraybackslash}p{0.3\columnwidth}YYY}
\toprule
Model & Rule & Obj-IoU & Func-IoU & F1-IoU \\
\midrule
\multirow{4}{*}{Qwen2-VL-2B} & TAM avg-subword & 26.66 & 69.56 & 38.55 \\
 & \cellcolor{lightgray}\method avg-subword & \cellcolor{lightgray}28.90 & \cellcolor{lightgray}90.08 & \cellcolor{lightgray}43.76 \\
 & TAM first-subword & 26.76 & 69.72 & 38.68 \\
 & \cellcolor{lightgray}\method first-subword & \cellcolor{lightgray}29.27 & \cellcolor{lightgray}89.89 & \cellcolor{lightgray}44.16 \\
\midrule
\multirow{4}{*}{InternVL2.5-2B} & TAM avg-subword & 21.20 & 63.94 & 31.84 \\
 & \cellcolor{lightgray}\method avg-subword & \cellcolor{lightgray}29.97 & \cellcolor{lightgray}75.90 & \cellcolor{lightgray}42.97 \\
 & TAM first-subword & 21.55 & 64.01 & 32.25 \\
 & \cellcolor{lightgray}\method first-subword & \cellcolor{lightgray}30.18 & \cellcolor{lightgray}75.93 & \cellcolor{lightgray}43.19 \\
\bottomrule
\end{tabularx}

\end{table}

\textbf{Component contribution.}
Table~\ref{tab:ablation} asks whether ER and PCR are both necessary. In this ablation, Baseline uses one evidence view without context subtraction, ER only applies view recomposition without PCR, and PCR only applies context residualization to a single evidence view ($M=1$). Mean sub. uses a uniform average of preceding-token maps, whereas TAM-style follows TAM by weighting these maps with text activation scores. All subtraction variants use the same fitted subtraction, positive clipping, and rank Gaussian filtering. On Qwen2-VL-2B, the baseline without ER or PCR obtains 31.57 F1-IoU; ER only and PCR only improve it to 36.65 and 41.78, respectively, while their combination reaches 44.45. The same ordering holds on LLaVA1.5-7B, where the corresponding scores are 31.00, 35.27, 42.19, and 45.34. This consistent pattern supports the complementarity of ER and PCR.

PCR only also outperforms both subtraction controls on both models. Mean recomposition with PCR outperforms max recomposition with PCR, supporting consensus aggregation rather than a single-view peak. Fig.~\ref{fig:ablation_visual} qualitatively illustrates the complementary effects of evidence aggregation and context residualization.

\textbf{Subword variants.}
Table~\ref{tab:oracle_free_subword} reports deterministic avg-subword and first-subword scoring variants. Under both rules, \method improves Obj-IoU, Func-IoU, and F1-IoU over TAM on Qwen2-VL-2B and InternVL2.5-2B. The F1-IoU gains are 5.21 and 5.48 points on Qwen2-VL-2B, and 11.13 and 10.94 points on InternVL2.5-2B, for avg-subword and first-subword scoring, respectively. The similar results under the two rules suggest that the gains are not tied to one deterministic treatment of subword tokens.

\textbf{Targeted PCR suppression.}
Table~\ref{tab:pcr_behavior} examines how PCR changes attribution mass across token categories. Compared with ER-only maps, PCR reduces total map mass more for function tokens and other tokens with low visual relevance than for object tokens, and the high-score area follows the same pattern. The mask-based In Ret. and Out Red. diagnostics are defined only for object tokens with matched masks, so they are not applicable to function tokens and other tokens with low visual relevance. For object tokens with matched masks, PCR retains 42.28\% of inside-mask evidence while reducing 82.97\% of outside-mask mass, consistent with stronger suppression outside the target object mask.

\begin{table}[!t]
\centering
\footnotesize
\setlength{\tabcolsep}{2.1pt}
\renewcommand{\arraystretch}{1.08}
\caption{PCR suppression behavior by token type on COCO Caption.}
\label{tab:pcr_behavior}
\begin{tabularx}{\columnwidth}{lYYYYY}
\toprule
Token group & $N$ & Supp. & Area Red. & In Ret. & Out Red. \\
\midrule
Object tokens & 479 & 78.81 & 88.09 & 42.28 & 82.97 \\
Function tokens & 1068 & 87.73 & 99.55 & N/A & N/A \\
Other tokens & 2453 & 85.39 & 95.36 & N/A & N/A \\
\bottomrule
\end{tabularx}

\end{table}

\textbf{Mechanism diagnostics.}
The component and aggregation ablations are consistent with the complementary roles of ER and PCR described in Sec.~\ref{subsec:bottleneck}. For ER, the advantage of mean over max recomposition is consistent with the view-consensus objective in Eq.~\eqref{eq:er_consensus}: averaging favors evidence repeatedly observed across views and reduces isolated peaks from a single view. The advantage of PCR over the two other subtraction controls is consistent with its context estimation for the current target: rank relevance determines which preceding-token maps contribute to the context map, while the fitted coefficient controls how much of this context is removed.

\begin{figure*}[!t]
    \centering
    \subfloat[Views, Qwen]{\includegraphics[width=0.245\linewidth]{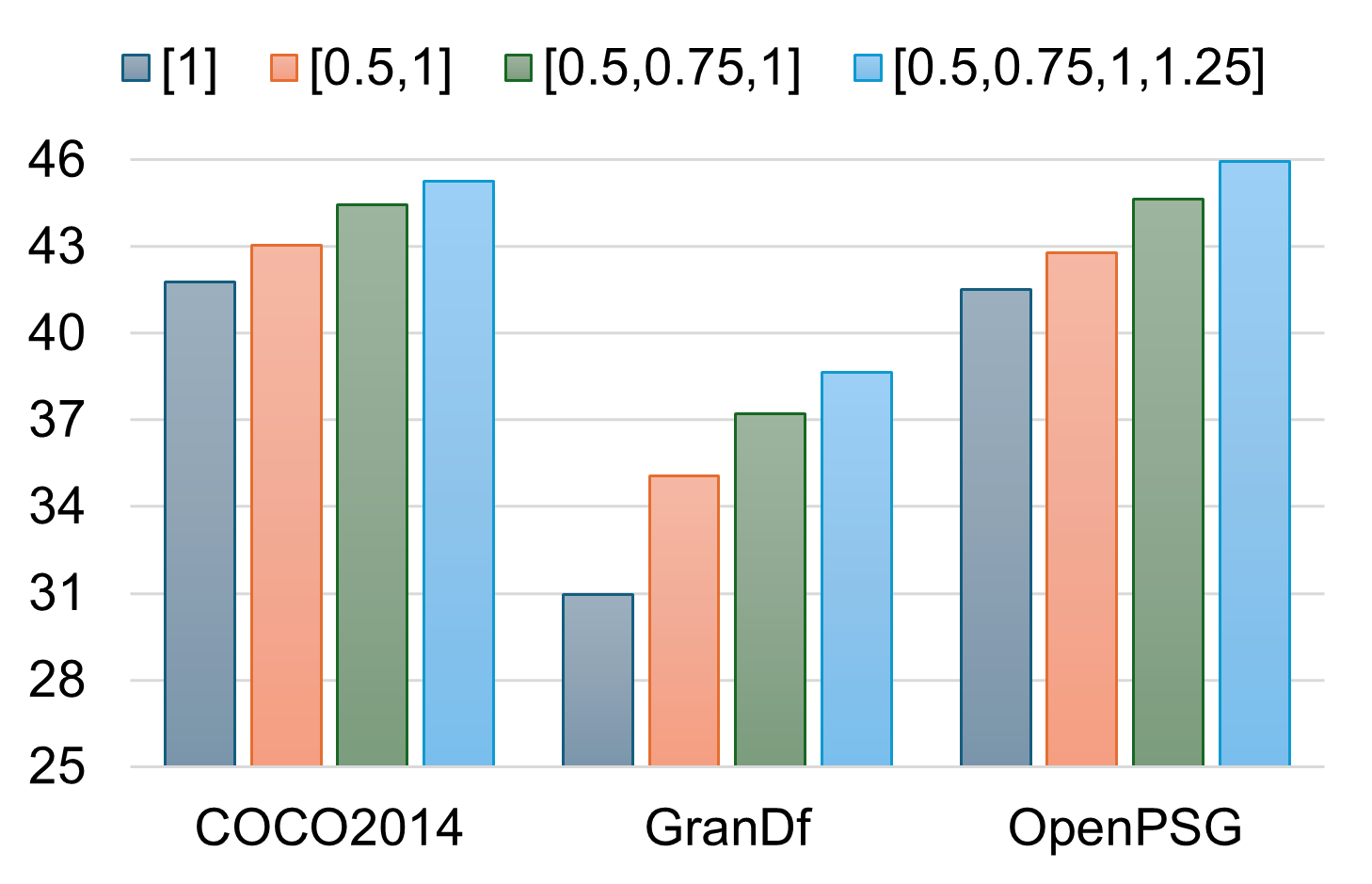}}\hfill
    \subfloat[Views, LLaVA]{\includegraphics[width=0.245\linewidth]{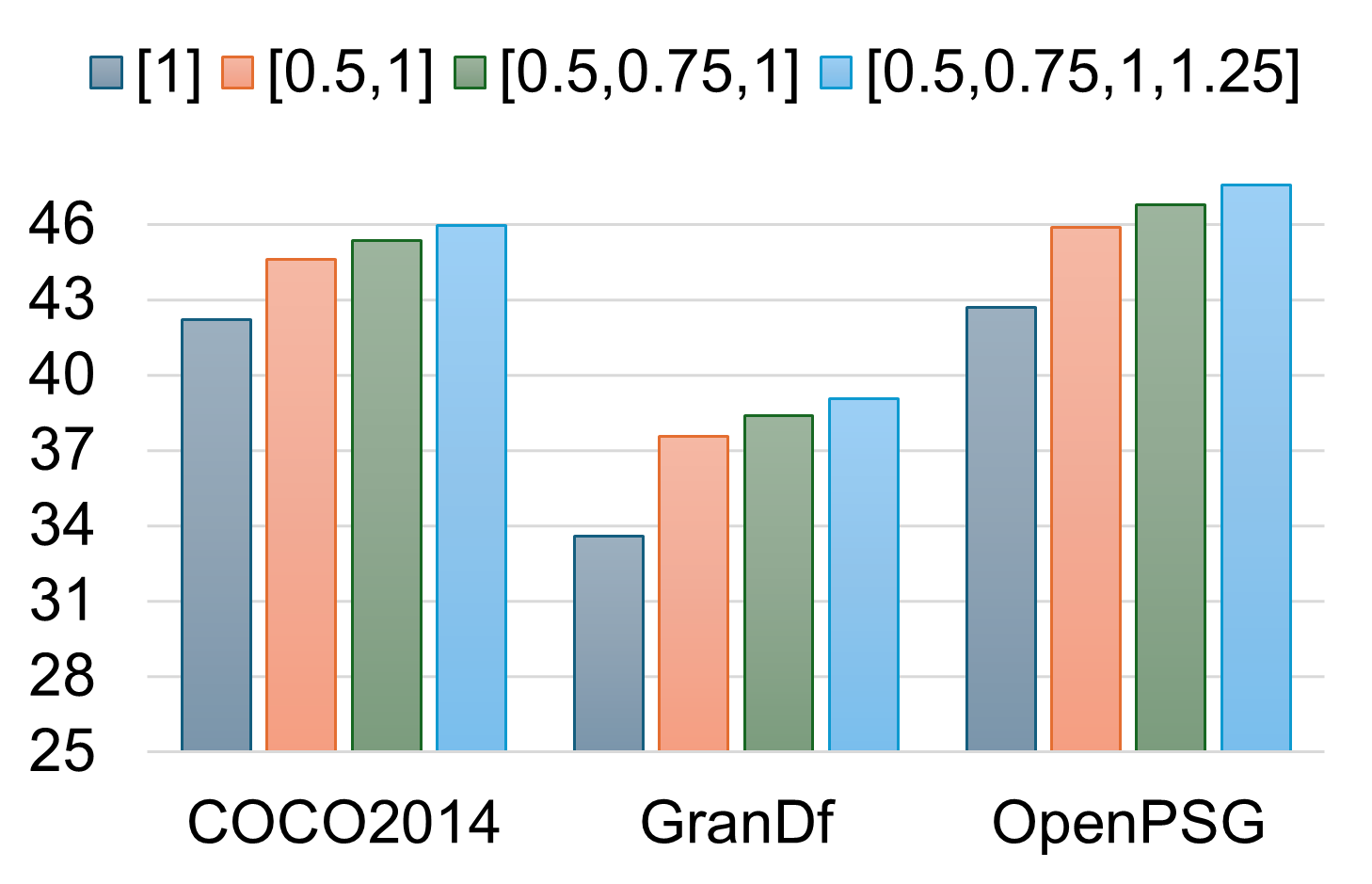}}\hfill
    \subfloat[Ranges, Qwen]{\includegraphics[width=0.245\linewidth]{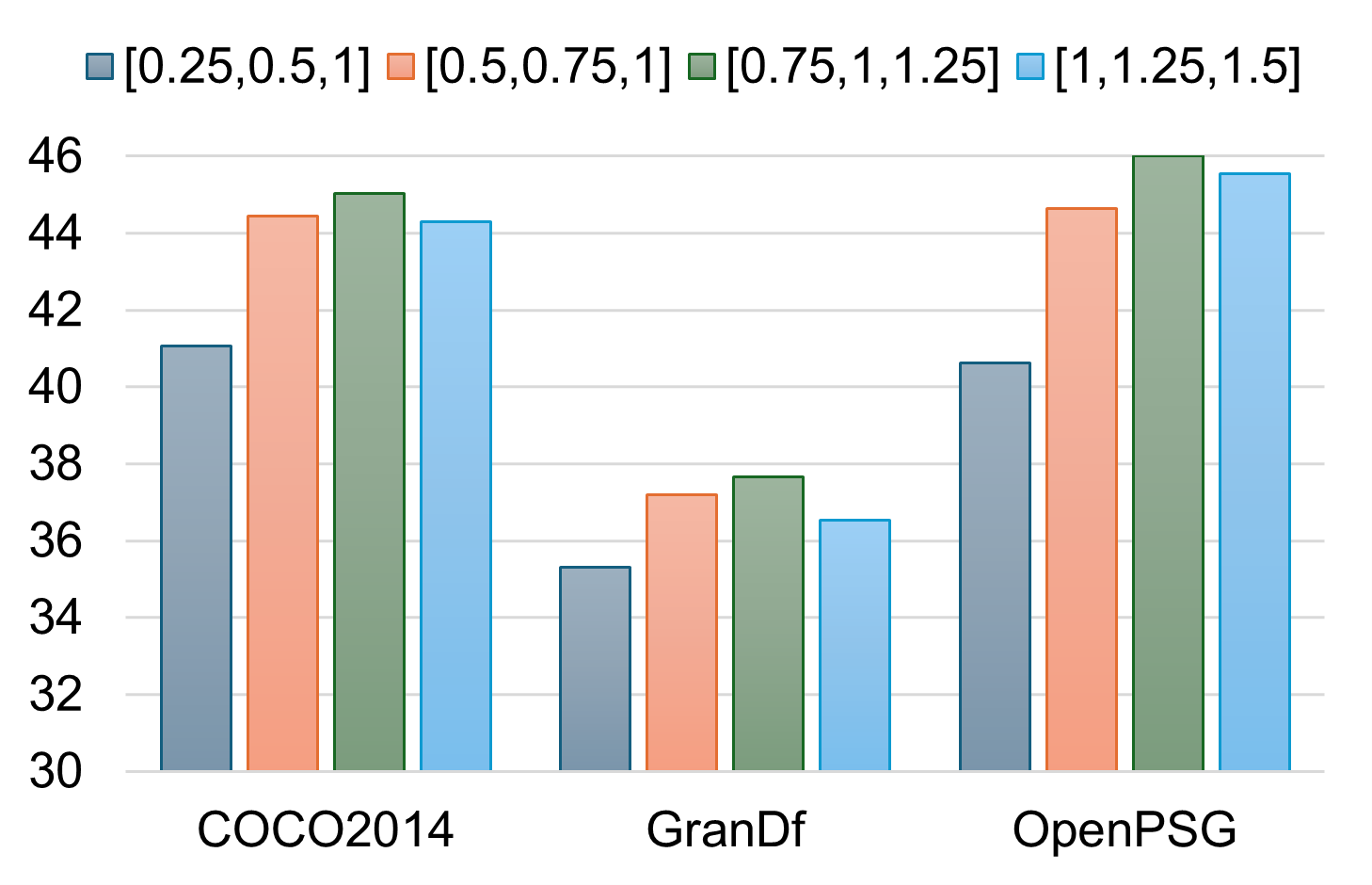}}\hfill
    \subfloat[Ranges, LLaVA]{\includegraphics[width=0.245\linewidth]{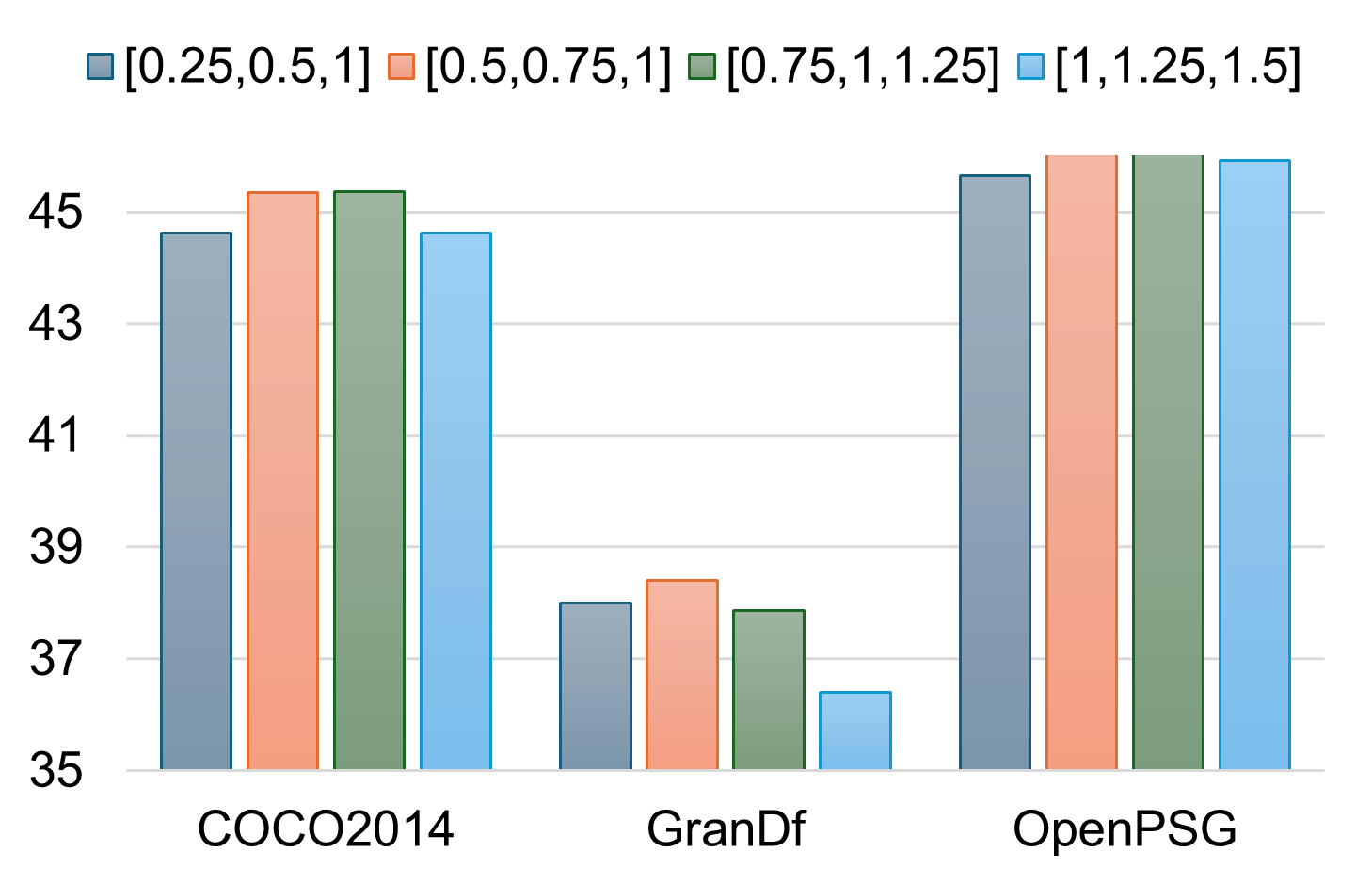}}
    \caption{Sensitivity to evidence-view number and scale range. The y-axis reports F1-IoU.}
    \label{fig:view_sensitivity}
\end{figure*}

\textbf{Evidence-view sensitivity.}
Fig.~\ref{fig:view_sensitivity} studies the number and scale range of ER evidence views. Moving from one evidence view to multiple views improves performance for both Qwen2-VL and LLaVA, with diminishing returns after a moderate number of views. This gain is consistent with reduced dependence on a single grid, while aggressive resizing can make view-specific evidence less stable. The default setting is therefore a practical accuracy-efficiency trade-off.

\begin{table}[!t]
\centering
\footnotesize
\setlength{\tabcolsep}{3.1pt}
\renewcommand{\arraystretch}{1.05}
\caption{Deletion and insertion faithfulness evaluation on COCO Caption at 10\%, 20\%, and 30\% evidence budgets.}
\label{tab:faithfulness}
\begin{tabularx}{\columnwidth}{ll|YYY|YYY}
\toprule
\multirow{2}{*}{Model} & \multirow{2}{*}{Method} &
\multicolumn{3}{c|}{Del. LogP Drop $\uparrow$} &
\multicolumn{3}{c}{Ins. LogP Gain $\uparrow$} \\
 & & 10\% & 20\% & 30\% & 10\% & 20\% & 30\% \\
\midrule
\multirow{3}{*}{Qwen2-VL-2B}
 & CAM & 0.871 & 1.540 & 1.964 & 1.786 & 2.451 & 2.736 \\
 & TAM & 0.942 & 1.587 & 2.182 & 2.000 & 2.575 & 2.792 \\
 & \method & \textbf{1.205} & \textbf{1.885} & \textbf{2.403} & \textbf{2.209} & \textbf{2.704} & \textbf{2.825} \\
\midrule
\multirow{3}{*}{InternVL2.5-2B}
 & CAM & 0.164 & 0.297 & 0.441 & 1.534 & 2.400 & 2.994 \\
 & TAM & 0.758 & 1.276 & 1.788 & 2.581 & 3.423 & 3.789 \\
 & \method & \textbf{1.043} & \textbf{1.678} & \textbf{2.201} & \textbf{2.968} & \textbf{3.619} & \textbf{3.903} \\
\midrule
\multirow{3}{*}{InternVL2.5-8B}
 & CAM & 0.123 & 0.287 & 0.480 & 1.224 & 2.294 & 2.914 \\
 & TAM & 1.067 & 1.630 & 2.023 & 2.787 & 3.401 & 3.658 \\
 & \method & \textbf{1.242} & \textbf{1.817} & \textbf{2.273} & \textbf{3.023} & \textbf{3.489} & \textbf{3.709} \\
\midrule
\multirow{3}{*}{InternVL3-2B}
 & CAM & 0.188 & 0.393 & 0.683 & 2.373 & 3.870 & 4.792 \\
 & TAM & 0.968 & 1.860 & 2.571 & 3.825 & 5.023 & 5.491 \\
 & \method & \textbf{1.260} & \textbf{2.192} & \textbf{2.941} & \textbf{4.436} & \textbf{5.358} & \textbf{5.659} \\
\bottomrule
\end{tabularx}

\end{table}

\subsection{Perturbation-Based Faithfulness Evaluation}
\label{subsec:faithfulness}

\begin{figure}[!t]
    \centering
    \includegraphics[width=\linewidth]{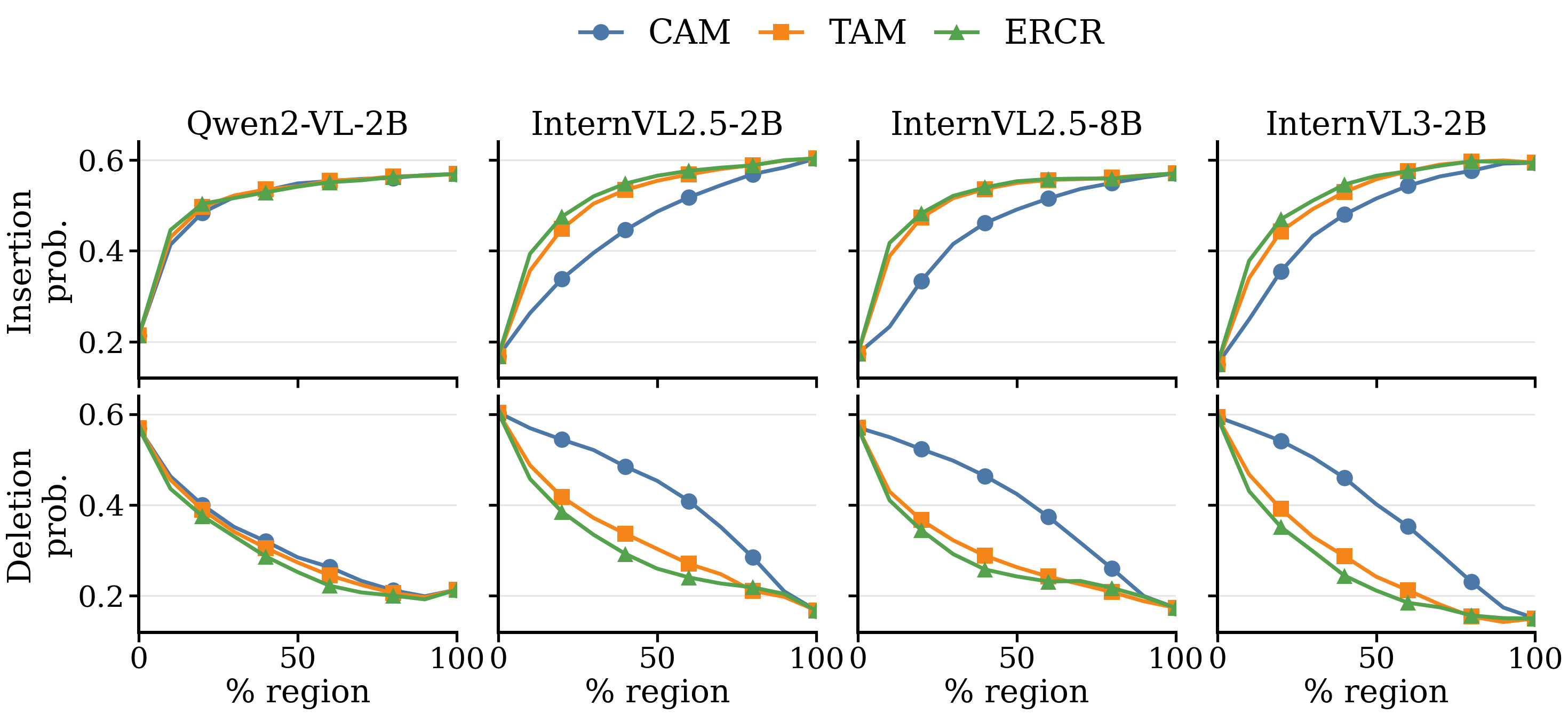}
    \caption{Target-token response trajectories under deletion and insertion on COCO Caption. Deletion removes top-ranked regions, while insertion restores them from a baseline image; lower deletion curves and higher insertion curves indicate more faithful evidence ranking.}
    \label{fig:faithfulness_curves}
\end{figure}

IoU-based plausibility metrics assess agreement with human masks, while fixed-prefix perturbations test whether highlighted regions affect the same target score in Eq.~\eqref{eq:target_score}. After constructing an attribution map, we conduct deletion and insertion evaluation~\cite{jacovi2020towards,fong2017interpretable,petsiuk2018rise,yeh2019infidelity} on object-matched COCO words. The generated response and prefix are fixed during rescoring, and the baseline image $B$ is the perturbed-image reference used for deletion and insertion. For a small image perturbation $\delta$, a first-order expansion gives
\begin{equation}
s_t(I+\delta)\approx s_t(I)+\left\langle\nabla_I s_t(I),\delta\right\rangle .
\label{eq:score_taylor}
\end{equation}
Low-budget deletion and insertion are therefore fixed-prefix faithfulness measurements: they ask whether top-ranked regions induce larger score changes before large masks, off-manifold baselines, or nonlinear saturation dominate. For a perturbation mask $M$ and baseline image $B$, define the fixed-prefix deletion score change
\begin{equation}
\Delta_t(M)=
s_t(I)-s_t\!\left((1-M)\odot I+M\odot B\right).
\label{eq:score_change}
\end{equation}
Accordingly, deletion and insertion are used as fixed-prefix faithfulness evaluation: they test whether top-ranked regions induce larger changes in the same target score. Let $M_q(\Atilde_t)$ denote the binary mask containing the top-$q$ fraction of pixels according to the attribution map. Deletion evaluates
\begin{equation}
\Delta_t^{\mathrm{del}}(q)=
s_t(I)-s_t\!\left((1-M_q)\odot I + M_q\odot B\right),
\label{eq:del_score}
\end{equation}
while insertion evaluates
\begin{equation}
\Delta_t^{\mathrm{ins}}(q)=
s_t\!\left(M_q\odot I + (1-M_q)\odot B\right)-s_t(B).
\label{eq:ins_score}
\end{equation}
This evaluation is related to infidelity-style explanation evaluation~\cite{yeh2019infidelity}: the generated response is kept fixed, and the target token is rescored by log-probability under the fixed prefix. Table~\ref{tab:faithfulness} reports low-budget deletion log-prob drop and insertion log-prob gain at 10\%, 20\%, and 30\%; Fig.~\ref{fig:faithfulness_curves} shows the full response trajectories.

\method is strongest at the reported low budgets on all four evaluated models. On Qwen2-VL-2B, the 30\% deletion log-prob drop improves from 2.182 for TAM to 2.403 for \method, while the insertion log-prob gain improves from 2.792 to 2.825. The margin is larger on InternVL3-2B, where \method reaches 2.941 deletion log-prob drop and 5.659 insertion log-prob gain at 30\%. These results indicate that the highlighted regions more reliably carry visual evidence for object words under the deletion and insertion protocol.

\begin{figure*}[!t]
    \centering
    \subfloat[Insertion]{
        \includegraphics[width=0.48\textwidth]{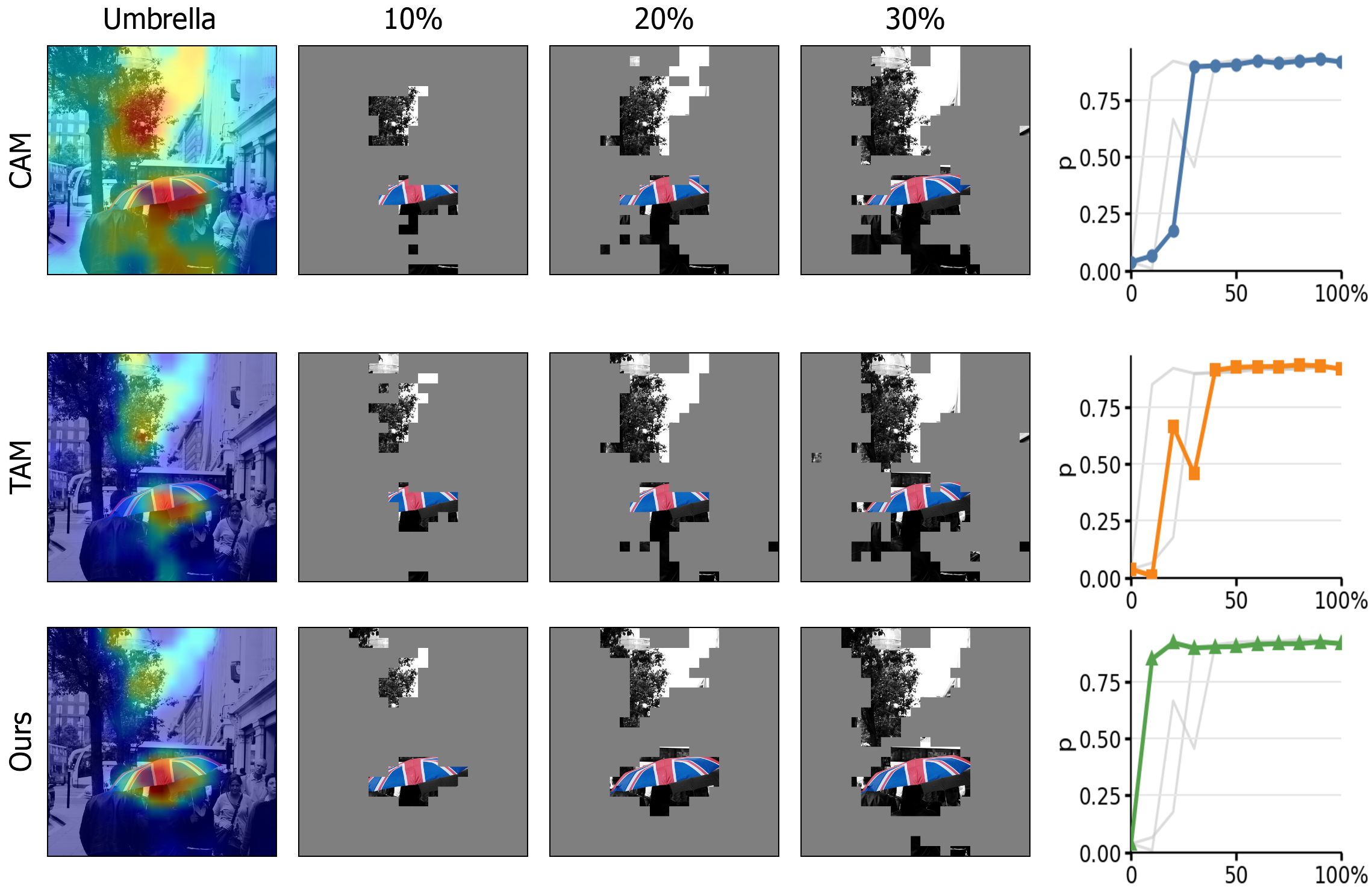}
    }\hfill
    \subfloat[Deletion]{
        \includegraphics[width=0.48\textwidth]{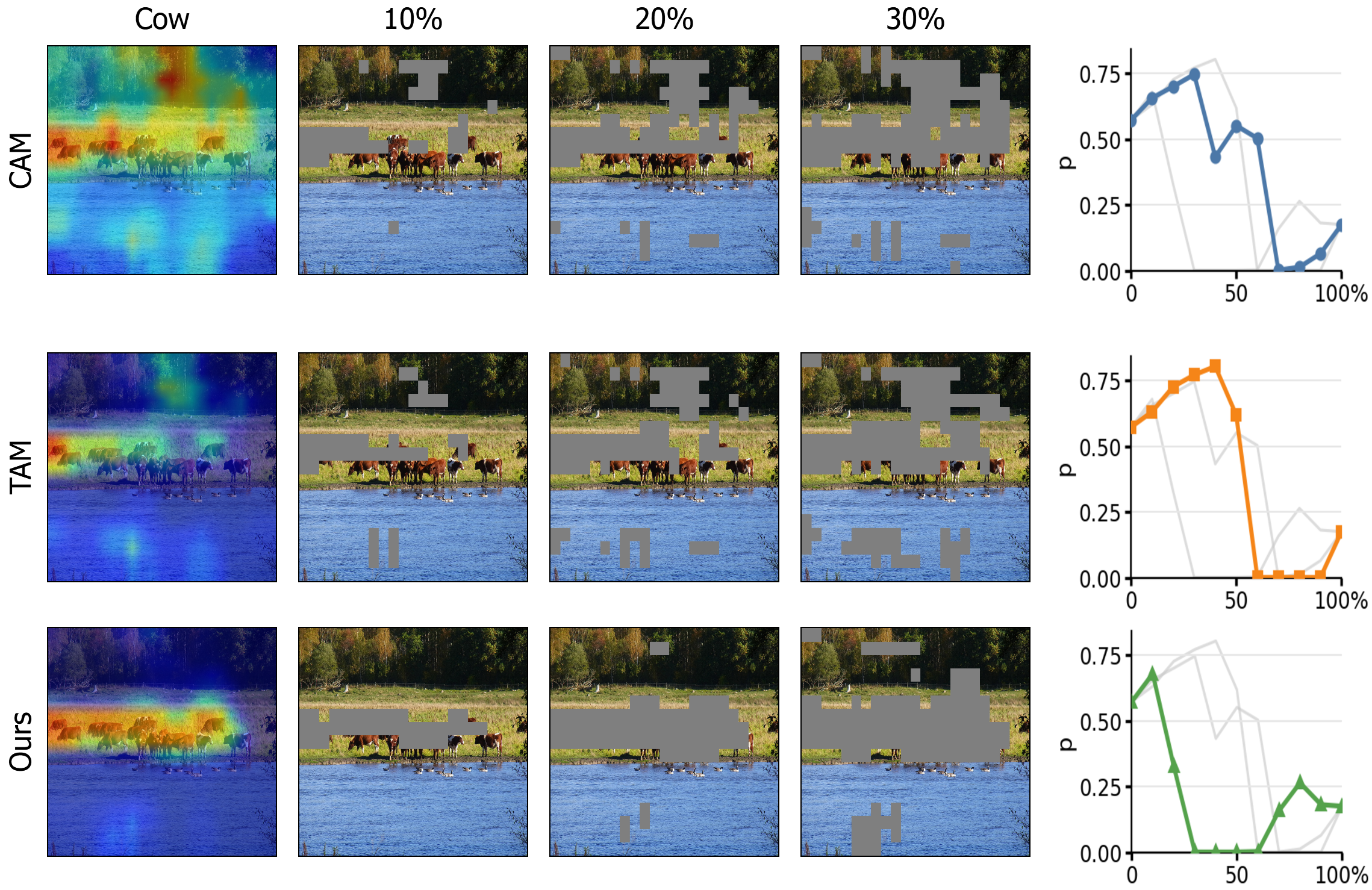}
    }
    \caption{Faithfulness demos on Qwen2-VL-2B. The insertion example restores top-ranked regions for ``umbrella'', while the deletion example removes top-ranked regions for ``cows''. Each method row shows the heatmap, three perturbation budgets, and the corresponding response curve.}
    \label{fig:faithfulness_demo_compact}
    \vspace{-0.2cm}
\end{figure*}

The response trajectories in Fig.~\ref{fig:faithfulness_curves} extend the low-budget observations across the full perturbation range. Lower deletion curves indicate that removing highlighted regions suppresses the target response earlier, while higher insertion curves indicate faster recovery when highlighted regions are restored. Fig.~\ref{fig:faithfulness_demo_compact} links these curves to concrete image regions in single examples.
\subsection{Efficiency Analysis}
\label{subsec:efficiency}

Table~\ref{tab:efficiency_qwen} reports average practical cost under the standard evaluation configuration. PCR adds little measured overhead because it mainly performs top-$k$ rank comparison and lightweight residualization. ER increases the number of evidence views from one to three, raising the average counted FLOPs from 3.53 to 6.51 TFLOPs. The full \method remains $1.20\times$ over TAM in measured runtime because auxiliary views require hidden-state readouts, with full decoding performed on the anchor view. Peak memory increases from 6.79 GB to 6.96 GB, making runtime the dominant practical cost.

\begin{figure*}[!t]
    \centering
    \subfloat[Qwen2-VL-2B]{
        \includegraphics[width=0.48\textwidth]{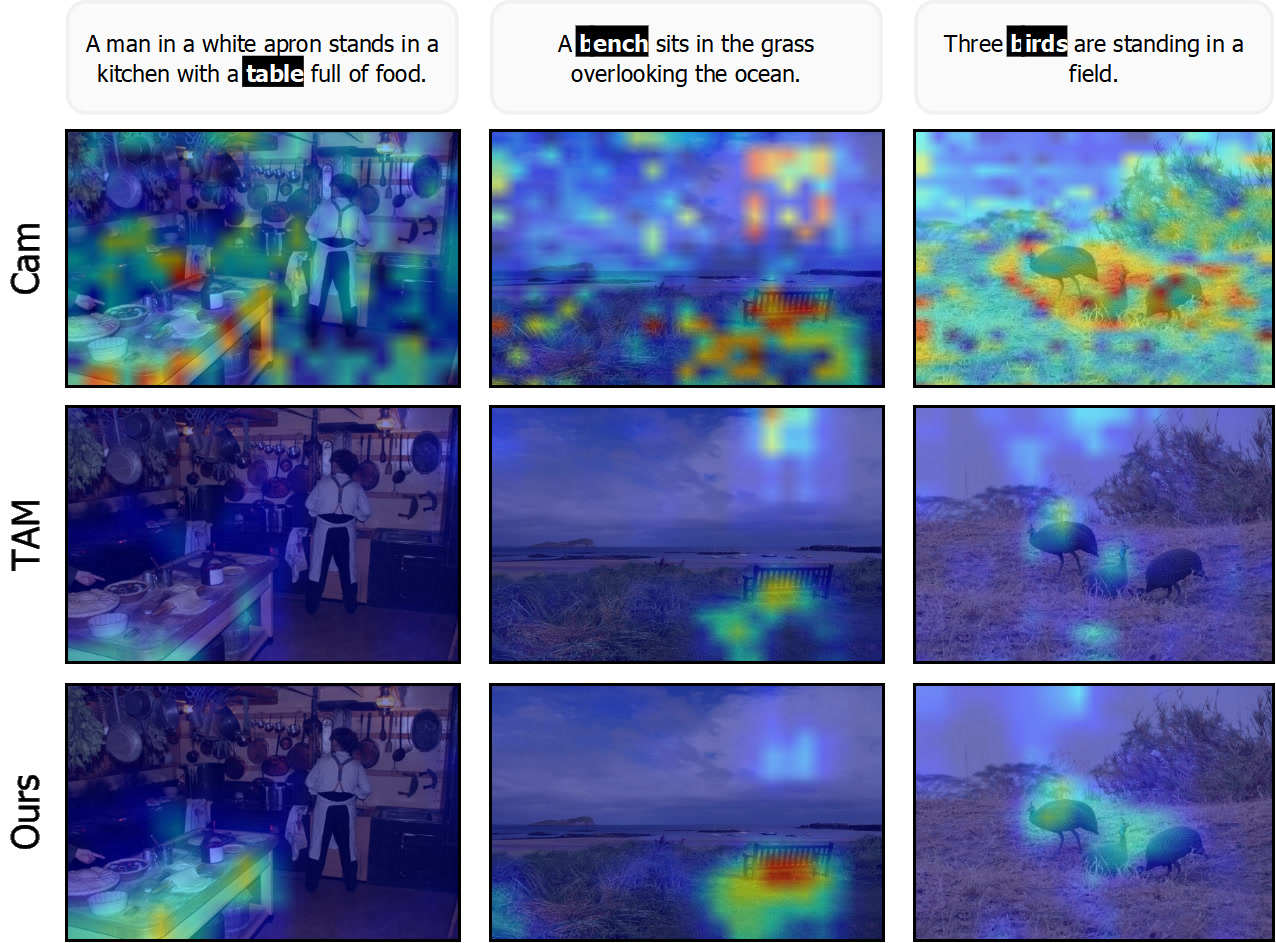}
        \label{fig:vis_qwen}}
    \hfill
    \subfloat[LLaVA-1.5-7B]{
        \includegraphics[width=0.48\textwidth]{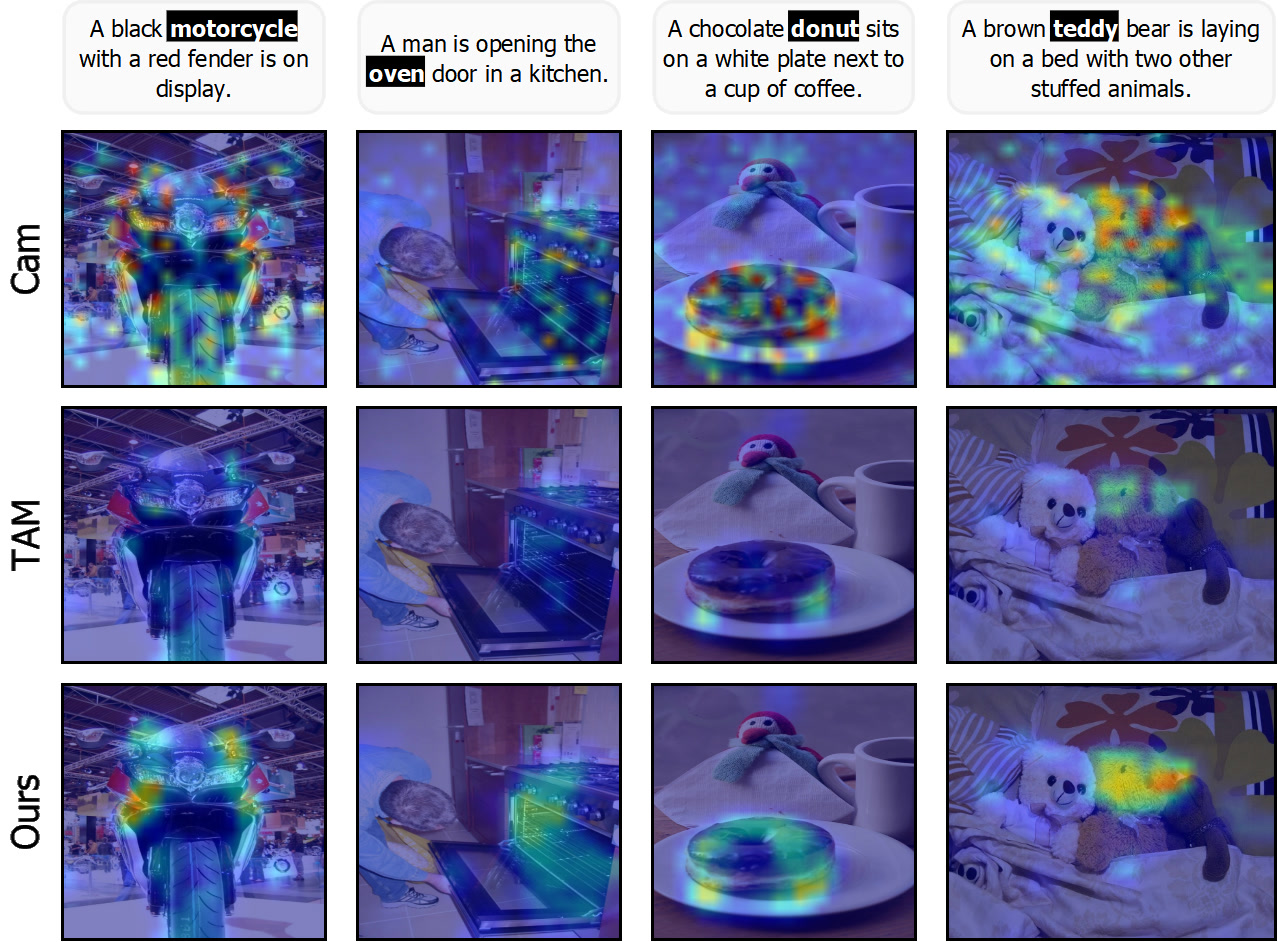}
        \label{fig:vis_llava}}
    \caption{Qualitative object-token attribution maps from Qwen2-VL-2B and LLaVA-1.5-7B. Warmer colors indicate higher target-token evidence.}
    \label{fig:vis_comparison}
    \vspace{-0.2cm}
\end{figure*}

\begin{table}[!t]
\centering
\footnotesize
\setlength{\tabcolsep}{4.0pt}
\renewcommand{\arraystretch}{1.08}
\caption{Runtime, peak memory, forward passes, and counted FLOPs on Qwen2-VL-2B with COCO Caption using one A100 80GB GPU.}
\label{tab:efficiency_qwen}
\begin{tabularx}{\columnwidth}{lYYYYY}
\toprule
Method & TFLOPs & Mem. (GB) & Fwd. & \makecell{Sec./\\Sample} & Overhead \\
\midrule
TAM & 3.53 & 6.79 & 1 & 1.61 & 1.00$\times$ \\
PCR only & 3.53 & 6.79 & 1 & 1.54 & 0.96$\times$ \\
ER only & 6.51 & 6.96 & 3 & 1.76 & 1.10$\times$ \\
\method & 6.51 & 6.96 & 3 & 1.93 & 1.20$\times$ \\
\bottomrule
\end{tabularx}

\vspace{-0.1cm}
\end{table}

\subsection{Qualitative Visualization}
\label{subsec:visualization}

% \FloatBarrier

\begin{figure*}[!t]
    \centering
    \subfloat[InternVL2.5-4B] {
        \includegraphics[width=0.428\textwidth]{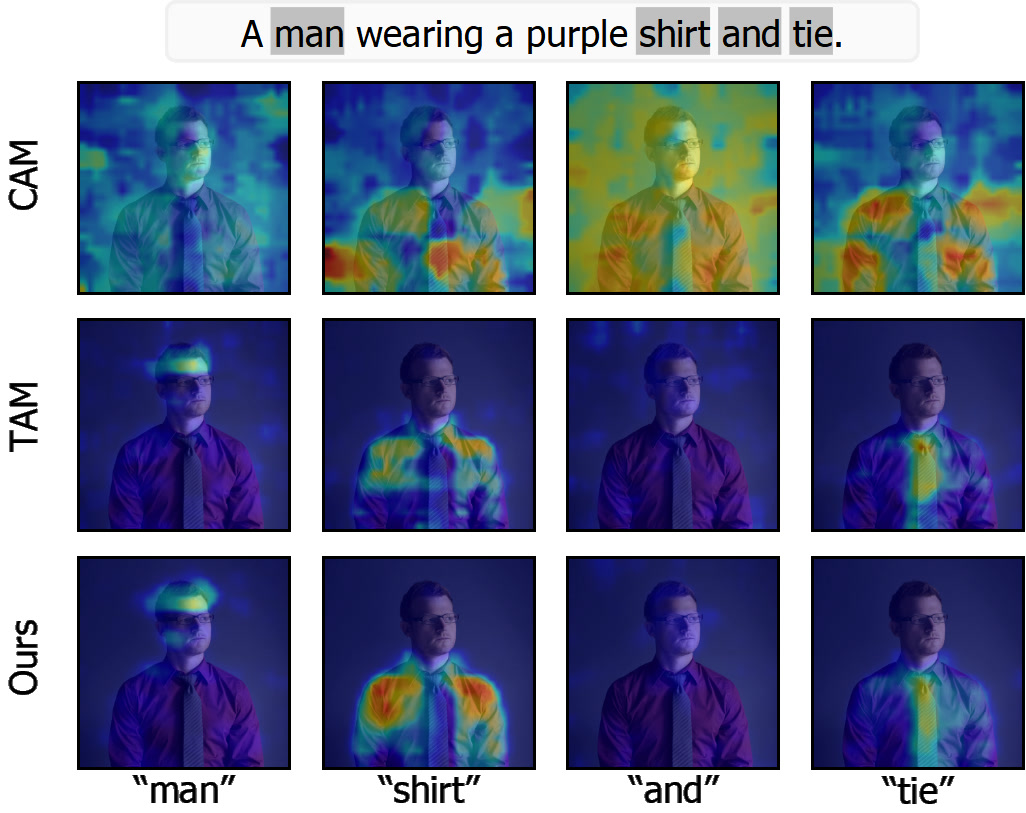}
        \label{fig:vis_internvl25_4b}}
    \hfill
    \subfloat[InternVL3.5-2B] {
        \includegraphics[width=0.532\textwidth]{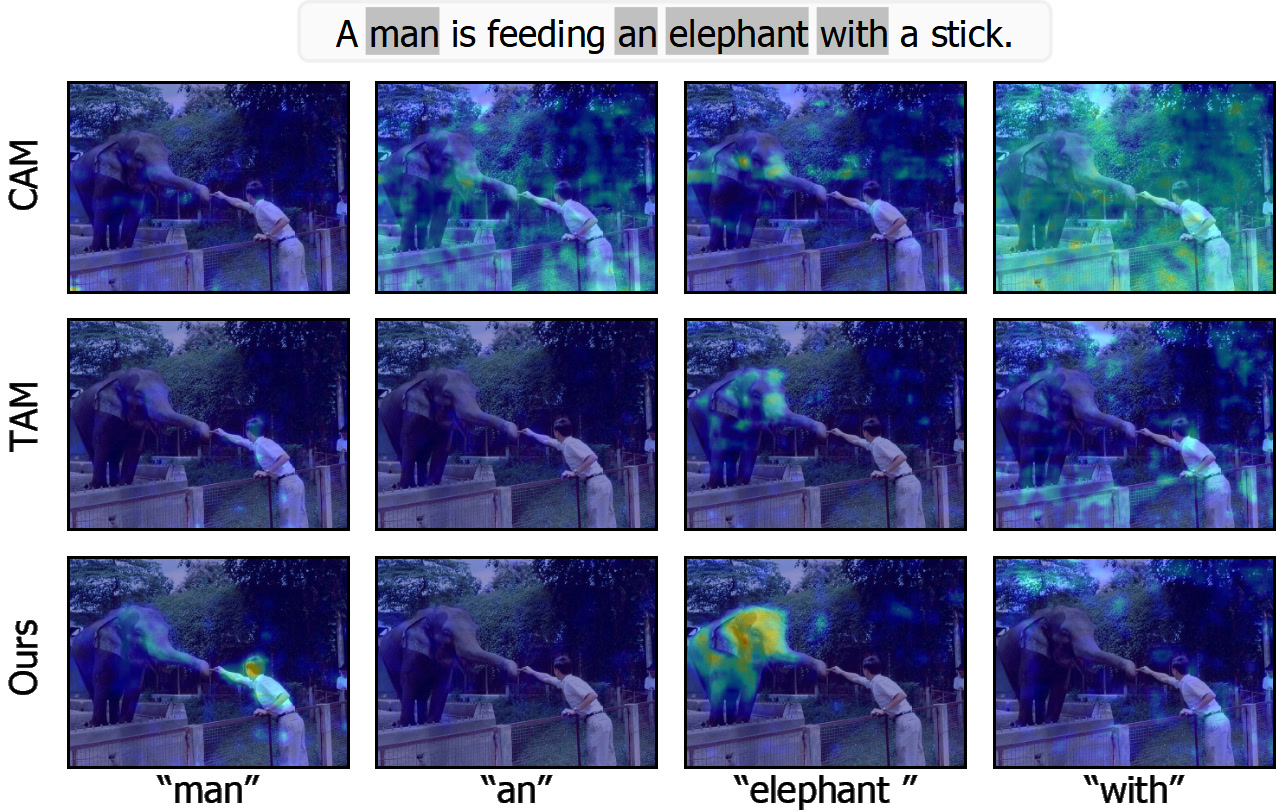}
        \label{fig:vis_internvl35_2b}}
    \caption{Token-specific attribution maps for a shared image within each InternVL model. Each panel visualizes multiple target tokens from one generated caption; warmer colors indicate higher target-token evidence.}
    \label{fig:vis_token_comparison}
\end{figure*}

Fig.~\ref{fig:vis_comparison} provides qualitative comparisons across different object-denoting target tokens and model families. \method produces less fragmented support for the current target and weaker diffuse context-associated activation, mirroring the Obj-IoU/F1 gains and PCR behavior trends. Fig.~\ref{fig:vis_token_comparison} further visualizes multiple target tokens from one generated caption for the same image within each InternVL model. Attribution shifts among the person, shirt, and tie in the InternVL2.5-4B example, and across the person, elephant, and their interaction region in the InternVL3.5-2B example, showing token-specific visual support within a shared scene.

\section{Discussion}
\label{sec:discussion}

\subsection{Why ER and PCR Help}
Visual tokens interact through the visual encoder and the multimodal transformer, so their hidden states already contain local and global context. The issue addressed by ER is the attribution readout: logit-lens attribution decodes each context-mixed visual token independently and assigns the score to a discrete token cell. ER mitigates this readout mismatch by aggregating evidence across views with different token-to-region assignments. As a view-consensus operation, it favors evidence that reappears across views and reduces reliance on a single tokenization grid for the same target token.

PCR separates context-map construction from spatial residualization. Rank relevance determines how preceding-token maps form the context map, while $\beta_t$ fits and subtracts the component represented in the current ER map. ER thus recomposes target evidence across views, whereas PCR refines the recomposed map using preceding-token context.

\subsection{Attribution Target and Evaluation}
The vocabulary coordinate $k_t$ specifies the attribution readout target, whereas $s_t(I)$ in Eq.~\eqref{eq:target_score} is the response-level score of $T_t$ conditioned on the anchor prefix $T_{<t}$. ER and PCR refine image-space attribution under this target definition without changing the logit-lens readout interface. We evaluate the resulting maps from two complementary perspectives: IoU-based plausibility metrics assess object-mask agreement and suppression of low-relevance visual evidence, while fixed-prefix deletion and insertion assess whether highlighted regions affect the same response score. The latter follows meaningful-perturbation and infidelity-style tests~\cite{fong2017interpretable,petsiuk2018rise,yeh2019infidelity} and examines whether the ranked regions carry visual evidence for that score.

\subsection{Scope and Practical Considerations}
The overhead of \method depends on the number and resolution of evidence views, and the default setting offers a practical accuracy-efficiency trade-off. When a preceding token and the current target share image support, the context map may also contain useful evidence; object-token retention diagnostics characterize this trade-off empirically. The local projection analysis assumes local linearity, a one-dimensional context direction, and bounded residuals. Future work may extend evaluation to relation and action tokens and adaptively weight evidence views.

\section{Conclusion}
\label{sec:conclusion}

We presented \method for token-level visual attribution in MLLMs. Evidence Recomposition reduces dependence on a single visual readout grid by aggregating evidence across multiple views for the same target token, while Predictive Context Residualization uses RBO-based rank relevance to mitigate preceding-token context interference. Experiments across several MLLM families and grounding-oriented benchmarks show that \method improves visual evidence for target tokens and mitigates preceding-token context interference under a TAM-compatible evaluation protocol. Overall, existing logit-lens attribution benefits from refining both the image-space readout and the preceding-token context component.

\bibliographystyle{IEEEtran}
\bibliography{refs}

\end{document}